\documentclass[10pt,twocolumn,letterpaper]{article}
\pdfoutput=1

\usepackage{iccv}
\usepackage{times}
\usepackage{epsfig}
\usepackage{graphicx}
\usepackage{amsmath}
\usepackage{amssymb}
\usepackage{amsmath}
\usepackage{amssymb}
\usepackage{dsfont}
\usepackage{bm}
\usepackage{xcolor,soul,colortbl}
\usepackage{array}
\usepackage{multirow}
\usepackage{subcaption}
\usepackage{dashrule}

\definecolor{CA}{HTML}{E6E6FA}
\definecolor{CS}{HTML}{FFEFD5}
\definecolor{CE}{HTML}{FFC0CB}

\definecolor{tcolor}{RGB}{219,50,54}
\definecolor{scolor}{RGB}{72,133,237}

\definecolor{col_rider}{RGB}{220,20,60}
\definecolor{col_veg}{RGB}{107,142,35}
\definecolor{col_ignore}{RGB}{0,0,0}
\definecolor{col_building}{RGB}{70,70,70}

\newcommand{\mm}[1]{\ensuremath{\bm{#1}}} 
\newcommand{\rtb}[1]{\rotatebox{90}{#1}}   

\newcolumntype{C}[1]{>{\centering\arraybackslash}p{#1}}	
\newcolumntype{L}[1]{>{\raggedright\arraybackslash}p{#1}}
\newcolumntype{R}[1]{>{\raggedleft\arraybackslash}p{#1}}

\newcommand{\src}[1]{#1_s}
\newcommand{\trg}[1]{#1_t}
\newcommand{\cX}{\mathcal{X}}

\newcommand{\cT}{\mathcal{T}}
\newcommand{\bx}{\mm x}

\newcommand{\seg}{\mathrm{seg}}
\newcommand{\dep}{\mathrm{dep}}
\newcommand{\adv}{\mathrm{adv}}

\usepackage{float}
\usepackage[export]{adjustbox}
\usepackage{subcaption}
\usepackage{graphics}
\usepackage{makecell}
\usepackage{multirow}
\usepackage{flushend}
\usepackage{xcolor,soul,colortbl}
\usepackage{authblk}

\usepackage[pagebackref=true,breaklinks=true,letterpaper=true,colorlinks,bookmarks=false]{hyperref}
\usepackage{pifont}
\iccvfinalcopy 

\begin{document}
	
	\title{DADA: Depth-Aware Domain Adaptation in Semantic Segmentation}
	\makeatletter
	\renewcommand\AB@affilsepx{\qquad\qquad \protect\Affilfont}
	\makeatother
	\author[1]{Tuan-Hung Vu}
	\author[1]{Himalaya Jain}
	\author[1]{Maxime Bucher}
	\author[1,2]{Matthieu Cord}
	\author[1]{Patrick P\'erez}
	\affil[1]{valeo.ai, Paris, France}
	\affil[2]{Sorbonne University, Paris, France}
	
	\twocolumn[{
		\renewcommand\twocolumn[1][]{#1}
		\maketitle
		\begin{center}
			\vspace{-0.8cm}
			\includegraphics[width=\linewidth]{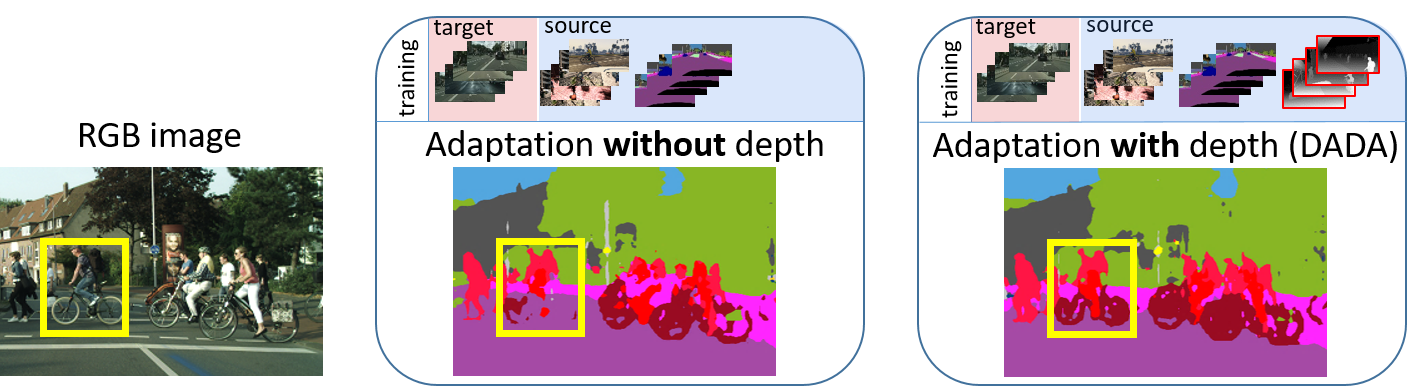}
			\captionof{figure}
			{\small
				\textbf{We propose a novel depth-aware domain adaptation framework (DADA) to efficiently leverage depth as privileged information in the unsupervised domain adaptation setting}. This example shows how semantic segmentation of a scene from the target domain benefits from the proposed approach, in comparison to state-of-the-art domain adaptation with no use of depth. In figure's top, we use different background colors (\textcolor{scolor}{blue} and \textcolor{tcolor}{red}) to represent \textcolor{scolor}{source} and \textcolor{tcolor}{target} information that are available during training. Here, annotated source domain data come from the synthetic SYNTHIA dataset and un-annotated target domain images are real scenes from Cityscapes. The cyclist highlighted by the yellow box is a good qualitative illustration of the improvement we obtain.
			}
			\label{fig:teaser}
		\end{center}
		\vspace{0.5cm}
	}]
	
	\begin{abstract}
		\vspace{-0.5cm}
		Unsupervised domain adaptation (UDA) is important for applications where large scale annotation of representative data is challenging. For semantic segmentation in particular, it helps deploy, on real ``target domain'' data, models that are trained on annotated images from a different ``source domain'', notably a virtual environment.
		To this end, most previous works consider semantic segmentation as the only mode of supervision for source domain data, while ignoring other, possibly available, information like depth.
		In this work, we aim at exploiting at best such a privileged information while training the UDA model.
		We propose a unified depth-aware UDA framework that leverages in several complementary ways the knowledge of dense depth in the source domain.  
		As a result, the performance of the trained semantic segmentation model on the target domain is boosted. Our novel approach indeed achieves state-of-the-art performance on different challenging synthetic-2-real benchmarks. Code and models are available at~\url{https://github.com/valeoai/DADA}.
	\end{abstract}
	
	\section{Introduction}
	Advances in deep convolutional neural networks (CNNs) brought significant leaps forward in many recognition tasks including semantic segmentation.
Still, predicting semantic labels for all imagery pixels is a challenging problem, especially when models are trained on one domain, named as \textit{source}, yet evaluated on another domain, named as \textit{target}.
The so-called \textit{domain gap} between source and target distributions often causes drastic drops in target performance.
Instead, autonomous critical systems such as self-driving cars require robust performance under diverse testing conditions, despite the lack of ubiquitous training data. For such systems, semantic segmentation models trained on sunny urban images taken in Roma should yield good results even on foggy scenes in London.

Techniques addressing the domain gap problem are usually classified as \textit{domain adaptation} (DA) \cite{csurka2017domain}.
In previous works, most DA settings are \textit{unsupervised} on the target side, \ie, only un-annotated target samples are available during the supervised training on source domain. This is referred to as unsupervised domain adaptation (UDA). In recent years, considerable progress has been made in UDA with approaches such as distribution discrepancy minimization with MMD \cite{long2015learning, yan2017mind} or adversarial training \cite{hoffman2016fcns, tsai2018learning}, generative approaches \cite{hoffman18a} and others.

In this UDA context, the \textit{synthetic-2-real} scenario, where source and target samples are synthetic and real images respectively, is especially appealing thanks to the zero-cost source label acquisition.
Moreover, recent virtual engines can simulate other sensory outputs like dense/sparse depth or radar; for example, the SYNTHIA dataset~\cite{Ros_2016_CVPR} provides corresponding virtual depth maps of synthesized urban scenes.
Most previous UDA works ignore such extra information, except for Lee \etal \cite{lee2018spigan} who propose to use depth for regularizing a style-transfer network.

In this work, we propose a new scheme to leverage depth information available in source domain for UDA.
We hypothesize that introducing additional depth-specific adaptation brings complementary effects to further bridge the performance gap between source and target at test time.
Toward this end,  we transform the segmentation backbone such that the depth information is embedded into a dedicated deep architecture by means of an auxiliary depth regression task.
Depth, operating as an additional source-domain supervision in our framework (only available while training), will be considered as a \textit{privileged information}.

Another challenge is to incorporate efficiently depth signals into the UDA learning. Addressing this concern, we introduce a new depth-aware adversarial training protocol based on the fusion of the network outputs.
Such a late fusion was inspired by our intuition that visual information at different depth levels should be treated differently.

The proposed approach is illustrated in Figure~\ref{fig:teaser}, where the benefit of depth-aware adaptation on key object categories like `human' and `vehicle' is visible. The contributions of this approach are the following: 
\begin{itemize}
	\setlength{\parskip}{0pt}
	\setlength{\itemsep}{1pt}
	\item Depth-aware UDA learning strategy: we introduce a novel depth-aware adaptation scheme, coined DADA learning, which simultaneously aligns segmentation-based and depth-based information of source and target while being aware of scene geometry.
	\item Depth-aware deep architecture: we propose a novel depth-aware segmentation pipeline, named DADA architecture, in which depth-specific and standard CNN appearance features are fused before being forwarded through the segmentation classifiers.
	\item State-of-the-art performance: evaluations show that our framework achieves SotA results on challenging \textit{synthetic-2-real} benchmarks. We also report ablation studies to provide insights into the proposed approach.
\end{itemize}
	
	\section{Related works}
	Unsupervised domain adaption has received a lot of attention in last few years \cite{csurka2017domain, ganin2015unsupervised, hoffman2016fcns, long2016unsupervised, tzeng2017adversarial, yan2017mind}. It is usually approached with domain discrepancy minimization, generative approaches or using some privileged information to guide the learning. Since we are only concerned with visual semantic segmentation in this work, we limit our review of UDA to approaches that aim at this task as well.

Various approaches of UDA for segmentation employ adversarial training to minimize cross-domain discrepancy. The main idea, stemming from generative adversarial networks (GANs) \cite{goodfellow2014generative}, is to train a discriminator for predicting the domain of the data (source or target) while the segmentation network tries to fool it (along with the supervised segmentation task on the source). Under the competition with the discriminator, the segmentation network tries to map its input to domain-agnostic  intermediate or final representation (which is the input to the discriminator), before accomplishing its task. This alignment with adversarial training is usually done in the feature space.
In \cite{hoffman2016fcns}, the feature alignment is done not only with adversarial training but also by transferring the label statistics of the source domain by category specific adaptation. \cite{chen2017no} uses adversarial training for class-level alignment on grid-wise soft pseudo-labels. In \cite{chen2018CVPR}, spatial-aware adaptation is conducted and a distillation loss is used to address specifically synthetic-to-real adaptation by enforcing the segmentation network's output to be similar to a reference network trained on real data. While most of these methods do alignment on the features space, recently \cite{tsai2018learning, vu2018advent} propose alignment on the output space. \cite{tsai2018learning} does the alignment on the prediction of the segmentation network and \cite{vu2018advent} proposes to do it on the \textit{weighted self-information} of the prediction probability.
\cite{tsai2019domain} and \cite{luo2018taking} extend the approach of \cite{tsai2018learning} by patch-level alignment and category-level adversarial loss respectively.
Another use of adversarial training for UDA is proposed in \cite{saito2017adversarial, saito2017maximum}, where the discrepancy between two instances of the same input from target domain is minimized while the classification layer tries to maximize it. The approach we propose also uses adversarial training but it takes this further by introducing the depth.

Another strategy that received much attention is the use of generative networks to turn source domain samples into target-like images. CyCADA \cite{hoffman18a} uses Cycle-GAN \cite{zhu2017unpaired} to generate target-like images conditioned on the source images, \ie, generated images contain the structure or semantic content of the source with the ``style'' (colors and textures) of the target domain. These generated images inherit the ground truth semantic segmentation of the conditioning source images and can then be used for supervised training of the segmentation network. 
\cite{zhu2018ECCV} aligns the source and target embeddings using GAN \cite{goodfellow2014generative} and replaces the cross-entropy loss by a \textit{conservative loss} (CL) which penalizes the easy and hard source examples. In DCAN \cite{wu2018dcan}, a similar generative approach is used with channel-wise feature alignment in the generator and segmentation networks. 

Some other interesting works have investigated combination of adversarial and generative approaches~\cite{murez2018CVPR, Sankaranarayanan_2018_CVPR, zhang2018CVPR}, self-training \cite{zou2018unsupervised}  and curriculum style learning~\cite{zhang2017curriculum}.

In order to help domain adaptation in various tasks, several works explore the use of \textit{privileged information} on the source data~\cite{chen2014recognizing, li2014exploiting, sarafianos2017adaptive}.
The idea of using privileged information (PI) for learning was first formulated by Vapnik \& Vashist (2009)~\cite{vapnik2009new}. PI is an additional information available only at training time. This is conceptually similar to humans learning new notions or concepts with the help of teacher's comment or explanation. Following~\cite{vapnik2009new} many works~\cite{hoffman2016learning, lopez2015unifying, mordan2018revisiting, Sharmanska_2013_ICCV, wang2015classifier} investigated PI for various tasks.
Recently, SPIGAN~\cite{lee2018spigan} has proposed such a UDA approach for semantic segmentation.
SPIGAN first uses a generative network to convert the source images into target-type images.
These new images are then used to train the segmentation network along with a depth regression network (privileged information network) in a supervised manner, thanks to the ground truth of the source images.
This work shows that the additional depth regression task helps in pixel level adaptation or, in other words, better captures the content of source images in the generated target-like images. 

In the present work, we also leverage a depth regression task to aid domain adaptation. However, this is accomplished in a way that notably differs form SPIGAN (which uses depth only as a regularization for the generator).

\begin{figure*}
	\centering
	\includegraphics[width=\textwidth, trim={0 0 0 0}, clip]{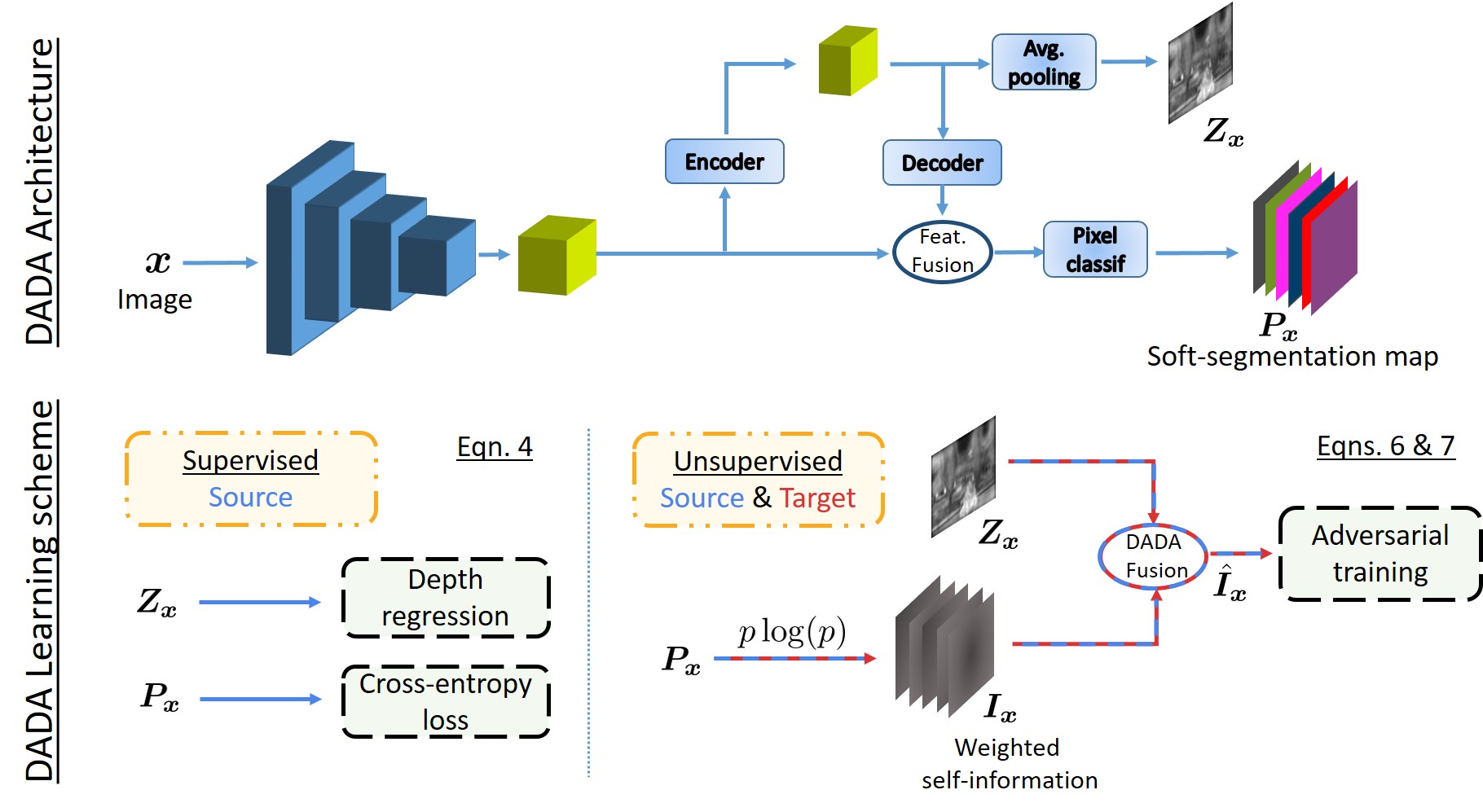}
	\vspace{-0.3cm}
	\caption{\small \textbf{DADA architecture (top) and DADA learning scheme (bottom)}. In the top part, the dark-blue stack shows the backbone CNN network; light-blue boxes symbolize the network modules; and green blocks stand for output features. In the lower part, the arrows drawn in \textcolor{scolor}{blue} and \textcolor{tcolor}{red} differentiate network flows of \textcolor{scolor}{source} and \textcolor{tcolor}{target} samples respectively. For convenient reference, over the learning blocks -- illustrated by dashed boxes -- we indicate the corresponding equation numbers.}
	\vspace{-0.3cm}
	\label{fig:dada_arch}
\end{figure*}
	
	\section{Depth-aware domain adaptation}
	In this section, we describe our proposed UDA method for semantic segmentation using depth.
Our goal is to use depth as the privileged information in the UDA setting to improve segmentation performance on the target domain.

To this end, we modify a network of semantic segmentation by including a monocular depth regression part.
More specifically, we  design a deep architecture to embed the depth in a dedicated residual block as illustrated in Figure~\ref{fig:dada_arch} (`DADA architecture' in top part). In Section~\ref{sec:dada_arch}, we detail DADA network architecture and the supervised learning used on source dataset.
The second part concerns the learning scheme for such a UDA approach. To get the full benefit of the geometry information, we propose a depth-aware adversarial learning scheme.
We argue that for domain adaptation, particularly in the urban settings, objects closer to the ego-camera should be emphasized more.
Our framework, illustrated in Figure~\ref{fig:dada_arch} (`DADA learning' in bottom part), is detailed in Section~\ref{sec:dada_learning}.

\subsection{DADA Network Architecture.}\label{sec:dada_arch}

Starting from an existing semantic segmentation architecture, we insert additional modules (1) to predict monocular depth as additional output and (2) to feed the information exploited by this auxiliary task back to the main stream.
More specifically, we adopt the \textit{residual auxiliary block} that was recently introduced in~\cite{mordan2018revisiting} for detection.
It amounts to grafting a new branch to the backbone CNN.
In this branch, the backbone CNN features are consecutively fed into three encoding convolutional layers, followed by an average pooling layer to output depth map predictions.
On the residual path back to the main branch, the encoded features (before the depth pooling) are decoded by a convolutional layer and fused with the backbone features.
The top-part of Figure~\ref{fig:dada_arch} shows the proposed hybrid architecture, mixing the auxiliary block architecture with the backbone one.

Importantly, for the feature-level fusion, we adopt an element-wise product, indicated as ``Feat. Fusion'' in the top part of Figure~\ref{fig:dada_arch}. To produce segmentation predictions, we feed-forward the fused features through the remaining classification modules.

\vspace{-0.3cm}
\paragraph{Source domain supervised training.} Our model is trained with supervised segmentation and depth losses on source domain.
We consider a training set $\src{\mathcal{T}}\subset  \mathbb{R}^{H\times W\times 3} \times (1,C) ^{H\times W} \times  \mathbb{R}_{+} ^{H\times W}$ of source color images of size $H\times W$ along with pixel-level $C$-class segmentation and depth annotations. 
Let $DADA$ be the network which takes an image $\mm x$ and jointly predicts a $C$-dimensional ``soft-segmentation map'' $DADA_{\mathrm{seg}}(\bx)~=~\mm P_{\mm x}~=~\big[ \mm P_{\mm x}^{(h,w,c)} \big]_{h,w,c}$~\footnote{Seen as empirical probabilities that sum to one over the $C$ classes.} and a depth map $DADA_{\mathrm{depth}}(\bx)=\big[ \mm Z_{\mm x}^{(h,w)} \big]_{h,w}$.
Similar to~\cite{kendall2017multitask}, we adopt the inverse depth representation, \ie, depth attenuates when moving away from the camera.
The parameters $\theta_{\mathrm{DADA}}$ of $DADA$ are learned to minimize the segmentation and depth losses on source samples $(\src{\mm x}, \src{\mm y}, \src{\mm z}) \in \mathcal{T}_s$:
\begin{align}\label{eq:mtlloss}
\vspace{-0.06cm}
\mathcal{L}_{\seg}(\src{\mm x}, \src{\mm y})&=-\sum_{h=1}^H\sum_{w=1}^W\sum_{c=1}^{C} \src{\mm y}^{(h,w,c)} \log \mm P_{\src{\mm x}}^{(h,w,c)},\\
\mathcal{L}_{\dep}(\src{\mm x}, \src{\mm z})&=-\!\sum_{h=1}^H\sum_{w=1}^W \text{berHu}\big(\mm Z_{\src{\mm x}}^{(h,w)}\!-\!\src{\mm z}^{(h,w)}\big),
\vspace{-0.06cm}
\end{align}
with the reverse Huber loss defined as~\cite{laina2016deeper}:
\begin{equation}
\vspace{-0.06cm}
\text{berHu}(e_z)=\begin{cases}
\left|e_z\right|,& \text{if } \left|e_z\right|\leq c,\\
\frac{e_z^2+c^2}{2c}              & \text{otherwise},
\end{cases}
\vspace{-0.06cm}
\end{equation}
where $c$ is a positive threshold that we fix in practice to $\frac{1}{5}$ of the maximum depth residual.
Empirically, the berHu loss is favorable for the depth regression task: samples with larger residuals are penalized more by the $\ell_2$ term, while gradients of small-residual samples are more underlined with $\ell_1$.

Finally, our DADA optimization problem on source domain is formulated as:
\begin{equation}\label{eq:onlysourcemtl}
\vspace{-0.05cm}
\min_{\theta_{\mathrm{DADA}}} \frac{1}{\left|\cT_s\right|}
\sum_{\cT_s} \mathcal{L}_{\seg}(\src{\mm x}, \src{\mm y})
+ \lambda_{\dep}\mathcal{L}_{\dep}(\src{\mm x}, \src{\mm z}),
\vspace{-0.05cm}
\end{equation}
with $\lambda_{\dep}$ a weighting factor for depth regression.

\subsection{DADA Adversarial Learning Scheme}\label{sec:dada_learning}
For UDA in semantic segmentation, the key idea is to align the source and target domains so that a discriminator network cannot distinguish between the domains.
We follow here the recent strategies that align features at the output level~\cite{tsai2018learning,vu2018advent}, \textit{i.e.}, the soft segmentation map $\mm P_{\mm x}$ produced by the segmentation network $DADA_{\seg}$ on input image $\mm x$.

We hereby question the plausibility of adapting such a methodology for an auxiliary space, \ie, depth prediction $\mm Z_{\mm x}$ in the present work, in the hope that the performance of the main task is improved.
We hypothesize that aligning source and target distributions also in the depth space implicitly bridges the domain gaps of the shared lower-level CNN representations and should bring 
improvements to the main task on target domain.

In order to carry out such a strategy, we propose a joint alignment:  we first merge  both signals, then we feed-forward the fused features as input to a discriminator.  

We illustrate DADA adversarial learning scheme in the lower right part of Figure~\ref{fig:dada_arch}.
More precisely:
\begin{itemize}
	\setlength{\parskip}{0pt}
	\setlength{\itemsep}{1pt}
    \item  We compute weighted self-information (``surprisal") maps \cite{vu2018advent} $\mm{I}_{\mm x}\in[0,1]^{H\times W\times C}$ defined as: 
    \begin{equation}
    \mm{I}_{\mm x}^{(h,w,c)} = -\mm P_{\mm x}^{(h,w,c)}\cdot\log \mm P_{\mm x}^{(h,w,c)}.
    \end{equation}
	\item We fuse the weighted self-information $\mm I_{\mm x}$ with the depth prediction $\mm Z_{\mm x}$ to produce a depth-aware map $\hat{\mm I}_{\mm x}$.
	The fusion of $\mm Z_{\mm x}$ and $\mm I_{\mm x}$, which we refer to as \textit{DADA fusion}, is the element-wise product of  $\mm I_{\mm x}$ with $\mm {Z_{\mm x}}$.
	As here the inverse depth is used, such a multiplication implies stronger attention toward scene elements that are closer to the ego-camera.
	\item Then, we do the adversarial adaptation on $\hat{\mm I}_{\mm x}$.
\end{itemize}

The depth-aware map $\hat{\mm I}_{\mm x}$ carries 3D-structural and geometrical information which should be consistent across the domains, thus alignment on this space is beneficial to adaptation as we will see in the next section.

Formally, given $\trg{\mathcal{X}}$ the set of un-annotated images in the target domain, the discriminator $D$ is trained to distinguish source \textit{vs.} target outputs (labeled as `1' and `0' respectively) with the following classification objective minimization:
\begin{equation}\label{eq:objdis}
\min_{\theta_D} \frac{1}{|\cT_s|}\sum_{\cT_s}\mathcal{L}_{D}(\hat{\mm{I}}_{\src{\mm x}}, 1) + \frac{1}{|\cX_t|}\sum_{\trg{\mathcal{X}}}\mathcal{L}_{D}(\hat{\mm{I}}_{\trg{\mm x}}, 0),
\end{equation}
and the DADA network is updated using the ``fooling'' objective minimization:
\begin{equation}\label{eq:objadv}
\min_{\theta_{\mathrm{DADA}}} \frac{1}{|\cX_t|}\sum_{\trg{\mathcal{X}}} \mathcal{L}_{D}(\hat{\mm{I}}_{\trg{\mm x}}, 1).
\end{equation}

At each training iteration, we feed the network a mini-batch of two samples coming from source and target domains.
The two objectives in (\ref{eq:onlysourcemtl}) and (\ref{eq:objadv}) are jointly optimized with a weighting factor $\lambda_{\adv}$ used for the adversarial part.
Gradients of all losses are accumulated and then back-propagated to update the network.

\vspace{-0.3cm}\paragraph{Discussion.} The only method using depth as privileged information in the same task as ours is SPIGAN~\cite{lee2018spigan}.
While in DADA, with feature fusion and DADA fusion, we exploit depth to enhance the appearance features and improve the source-target alignment, SPIGAN leverages depth as a regularization for the pixel-level alignment generator.
We argue that our way of using depth is more explicit, which, in return, gets more benefit from the privileged information.
Moreover, thanks to the residual fusion in the auxiliary block presented in Section~\ref{sec:dada_arch}, depth signals are directly taken into account while deriving main task predictions.
Such a residual fusion spreads beneficial effects of the depth-specific adaptation to the main segmentation task.
	
	\section{Experiments}
	\begin{table*}[t!]
	{
		\begin{center}
			\begin{tabular}{@{}L{2.9cm}@{}|@{}C{0.8cm}@{}|@{}C{0.7cm}@{}C{0.7cm}@{}C{0.7cm}@{}C{0.7cm}@{}C{0.7cm}@{}C{0.7cm}@{}C{0.7cm}@{}C{0.7cm}@{}C{0.7cm}@{}C{0.7cm}@{}C{0.7cm}@{}C{0.7cm}@{}C{0.7cm}@{}C{0.7cm}@{}C{0.7cm}@{}C{0.7cm}@{}|@{}c@{}|@{}c@{}|@{}c@{}}
				\multicolumn{21}{c}{ \rule{0pt}{2.5ex} SYNTHIA $\rightarrow$ Cityscapes (16 classes)}\\
				\hline
				\hline
				Models & \rtb{Depth} & \rtb{road} & \rtb{sidewalk\,\,} & \rtb{building} & \rtb{wall*} & \rtb{fence*} & \rtb{pole*} & \rtb{light} & \rtb{sign} & \rtb{veg} & \rtb{sky} & \rtb{person} & \rtb{rider} & \rtb{car} & \rtb{bus} & \rtb{mbike} & \rtb{bike} & \,\,\,\,\,\rtb{mIoU}\,\,\,\,\, & \,\,\,\,\,\rtb{D-Gain}\,\,\,\,\, & \,\,\,\,\,\rtb{mIoU*}\,\,\,\,\,\\
				\hline
				\rule{0pt}{3ex}SPIGAN-no-PI~\cite{lee2018spigan}&&69.5&29.4&68.7&4.4&0.3&32.4&5.8&15.0&81.0&78.7&52.2&13.1&72.8&23.6&7.9&18.7&35.8&-&41.2\\
				\rule{0pt}{3ex}SPIGAN~\cite{lee2018spigan}&\checkmark&71.1&29.8&71.4&3.7&0.3&\textbf{33.2}&6.4&\textbf{15.6}&81.2&78.9&52.7&13.1&75.9&25.5&10.0&20.5&36.8&1.0&42.4\\
				\hline
				\rule{0pt}{3ex}AdaptSegnet~\cite{tsai2018learning}&&79.2&37.2&78.8&-&-&-&9.9&10.5&78.2&80.5&53.5&19.6&67.0&29.5&21.6&31.3&-&-&45.9\\
				\rule{0pt}{3ex}AdaptPatch~\cite{tsai2019domain}&&82.2&39.4&79.4&-&-&-&6.5&10.8&77.8&82.0&54.9&21.1&67.7&30.7&17.8&32.2&-&-&46.3\\
				\rule{0pt}{3ex}CLAN~\cite{luo2018taking}&&81.3&37.0&80.1&-&-&-&\textbf{16.1}&13.7&78.2&81.5&53.4&21.2&73.0&32.9&\textbf{22.6}&30.7&-&-&47.8\\
				\rule{0pt}{3ex}AdvEnt~\cite{vu2018advent}&&87.0&44.1&79.7&\textbf{9.6}&\textbf{0.6}&24.3&4.8&7.2&80.1&83.6&\textbf{56.4}&\textbf{23.7}&72.7&32.6&12.8&33.7&40.8&-&47.6\\
				\rowcolor[gray]{.92}[0pt][0pt]\rule{0pt}{3ex}DADA &\checkmark&\textbf{89.2}&\textbf{44.8}&\textbf{81.4}&6.8&0.3&26.2&8.6&11.1&\textbf{81.8}&\textbf{84.0}&54.7&19.3&\textbf{79.7}&\textbf{40.7}&14.0&\textbf{38.8}&\textbf{42.6}&\textbf{1.8}&\textbf{49.8}
			\end{tabular}
		\end{center}
	}
	\vspace{-0.3cm}
	\caption{\small \textbf{Semantic segmentation performance mIoU (\%) on Cityscapes validation set of different models trained on SYNTHIA}. Top and bottom sub-tables correspond to VGG-16-based and ResNet-101-based models respectively. For methods making use of depth, we report the absolute depth-driven mIoU gain (D-Gain). We also show the mIoU (\%) of the $13$ classes (mIoU*) excluding classes with *.}
	\vspace{-0.2cm}
	\label{lbl:tbl_res_16classes}
\end{table*}

\begin{table*}[t!]
	{
		\begin{center}
			\begin{tabular}{@{}C{0.8cm}@{}|@{}L{3.0cm}@{}|@{}C{0.8cm}@{}||@{}C{0.7cm}@{}C{0.7cm}@{}C{0.7cm}@{}C{0.7cm}@{}C{0.7cm}@{}C{0.7cm}@{}C{0.7cm}@{}|@{}c@{}||@{}C{0.7cm}@{}C{0.7cm}@{}C{0.7cm}@{}C{0.7cm}@{}C{0.7cm}@{}C{0.7cm}@{}C{0.7cm}@{}|@{}c@{}}
				\multicolumn{3}{l||}{}&\multicolumn{8}{c||}{\small \rule{0pt}{2.5ex} (a) SYNTHIA $\rightarrow$ Cityscapes (7 classes)}&\multicolumn{8}{c}{\small \rule{0pt}{2.5ex} (b) SYNTHIA $\rightarrow$ Vistas (7 classes)}\\
				\hline
				\hline
				\rtb{Reso.} & \centering Models & \rtb{Depth} & \rtb{flat} & \rtb{const.\,\,} & \rtb{object} & \rtb{nature} & \rtb{sky} & \rtb{human} & \rtb{vehicle\,\,} & \,\,\,\,\,\rtb{mIoU}\,\,\,\,\,& \rtb{flat} & \rtb{const.\,\,} & \rtb{object} & \rtb{nature} & \rtb{sky} & \rtb{human} & \rtb{vehicle\,\,} & \,\,\,\,\,\rtb{mIoU}\,\,\,\,\,\\
				\hline
				\multirow{4}{*}{\rtb{$320\times640$\,\,\,\,\,}}&\rule{0pt}{3ex}\,\,SPIGAN-no-PI~\cite{lee2018spigan}&&90.3&58.2&6.8&35.8&69.0&9.5&52.1&46.0&53.0&30.8&3.6&14.6&53.0&5.8&26.9&26.8\\
				&\rule{0pt}{3ex}\,\,SPIGAN~\cite{lee2018spigan}&\checkmark&\textbf{91.2}&66.4&9.6&56.8&71.5&17.7&60.3&53.4&74.1&47.1&6.8&43.3&83.7&11.2&42.2&44.1\\
				\cline{2-19}
				&\rule{0pt}{3ex}\,\,Advent*~\cite{vu2018advent}&&86.3&72.7&12.0&70.4&\textbf{81.2}&29.8&62.9&59.4&82.7&51.8&18.4&\textbf{67.8}&79.5&22.7&54.9&54.0\\
				&\rule{0pt}{3ex}\,\,DADA&\checkmark&89.6&\textbf{76.0}&\textbf{16.3}&\textbf{74.4}&78.3&\textbf{43.8}&\textbf{65.7}&\textbf{63.4}&\textbf{83.8}&\textbf{53.7}&\textbf{20.5}&62.1&\textbf{84.5}&\textbf{26.6}&\textbf{59.2}&\textbf{55.8}\\
				\hline
				\hline
				\multirow{3}{*}{\rtb{Full\,\,\,\,\,}}&\rule{0pt}{3ex}\,\,Advent*~\cite{vu2018advent}&&89.6&77.8&22.1&\textbf{76.3}&81.4&54.7&68.7&67.2&86.9&58.8&30.5&74.1&85.1&48.3&72.5&65.2\\
				&\rule{0pt}{3ex}\,\,DADA&\checkmark&\textbf{92.3}&\textbf{78.3}&\textbf{25.0}&75.5&\textbf{82.2}&\textbf{58.7}&\textbf{72.4}&\textbf{70.4}&\textbf{86.7}&\textbf{62.1}&\textbf{34.9}&\textbf{75.9}&\textbf{88.6}&\textbf{51.1}&\textbf{73.8}&\textbf{67.6}\\
				\cline{2-19}
				&\rule{0pt}{3ex}\,\,Oracle (only-target)&&97.6&87.9&46.0&87.9&88.8&69.1&88.6&80.8&95.0&84.2&54.8&87.7&97.2&70.2&87.5&82.4
			\end{tabular}
		\end{center}
	}
	\vspace{-0.3cm}
	\caption{\small \textbf{Semantic segmentation performance mIoU (\%) in 7-classes setups.} (a) Cityscapes and (b) Vistas validation set. We report results produced at different resolutions. AdvEnt* is the adaptation of AdvEnt published code to 7-classes set-ups.}
	\vspace{-0.3cm}
	\label{lbl:tbl_res_7classes}
\end{table*}

This section presents quantitative and qualitative results. 
We introduce in Section~\ref{sec:exp_details} the \textit{synthetic-2-real} benchmarks used in this work.
We then analyze the performance of the proposed model in Section~\ref{sec:exp_res} and report ablation studies in Section~\ref{sec:exp_abl}.

\subsection{Experimental details}
\label{sec:exp_details}

\paragraph{Datasets.}
In this work, we use SYNTHIA dataset~\cite{Ros_2016_CVPR} as the source domain.
It is composed of $9,400$ synthetic images annotated with pixel-wise semantic labels and depth.
Similar to previous works~\cite{lee2018spigan, tsai2018learning, vu2018advent}, we adopt the split SYNTHIA-RAND-CITYSCAPES using Cityscapes-style annotations.
For the target domain, we use either Cityscapes~\cite{cordts2016cityscapes} or Mapillary Vistas~\cite{MVD2017} datasets.
What follows are our experimental set-ups in detail:
\begin{itemize}
	\item \textit{SYNTHIA$\rightarrow$Cityscapes (16 classes)}: This is a standard evaluation protocol used in previous works. The models are trained on the 16 classes common to SYNTHIA and Cityscapes. Similar to~\cite{tsai2018learning, vu2018advent}, we also report performance on the $13$-class subset.
	\item \textit{SYNTHIA$\rightarrow$Cityscapes~/~Vistas (7 classes)}: Following~\cite{lee2018spigan}, we conduct experiments on the $7$ categories that are common to SYNTHIA, Cityscapes and Vistas.
\end{itemize}
\paragraph{Network architecture.}
In our experiments, we adopt Deeplab-V2~\cite{chen2018deeplab} based on ResNet-101~\cite{He2015} as the backbone segmentation architecture.
Like~\cite{tsai2018learning, vu2018advent}, we apply Atrous Spatial Pyramid Pooling (ASPP) with sampling rates of \{6, 12, 18, 24\}.
Segmentation prediction is only done on the \textit{conv5} features.
For the adversarial training, we use DCGAN's discriminator~\cite{radford2015unsupervised} composed of $4$ sequential convolutional layers with leaky-ReLUs as activation functions.

The encoding module used for depth regression has three consecutive convolutional layers: the first and last ones have kernel size of $1$; the middle layer has kernel size of $3$ with a suitable zero-padding to ensure the same input and output resolutions. Each layer uses $4$ times fewer channels than the previous one.
In the decoding part, we feed the encoded features through a $1\times1$ convolutional layer.
The decoding layer has the same number of output channels as the channel size of the ResNet-101 backbone feature.

\vspace{-0.3cm}
\paragraph{Implementation details.}
Implementations are done with the PyTorch deep learning framework~\cite{paszke2017automatic}.
To train and validate our models, we use a single NVIDIA 1080TI GPU with 11GB memory.
We initialize our models with the ResNet-101~\cite{He2015} pre-trained on the ImageNet dataset~\cite{deng2009imagenet}.
Segmentation and depth regression networks are trained by a standard Stochastic Gradient Descent optimizer~\cite{bottou2010large} with learning rate $2.5\times 10^{-4}$, momentum $0.9$ and weight decay $10^{-4}$. 
For discriminator training, we adopt Adam optimizer~\cite{kingma2014adam} with learning rate $10^{-4}$.
In all experiments, we fixed $\lambda_{\dep}$ as $10^{-3}$ for the depth regression task and used $\lambda_{\adv} = 10^{-3}$ to weight the adversarial loss.
\subsection{Results} \label{sec:exp_res}

We present results of the proposed DADA approach in comparison to different baselines.
On the three benchmarks, our models achieve state-of-the-art performance.
Our extensive study shows the benefit of leveraging depth as privileged information with our DADA framework for UDA in semantic segmentation.

\begin{figure*}[t!]
	\begin{center}
		\begin{subfigure}[t]{0.24\textwidth}\centering
			\caption{Input image}\vspace{-0.2cm}
			\includegraphics[width=.98\textwidth]{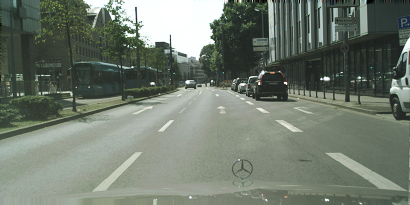}
		\end{subfigure}
		\begin{subfigure}[t]{0.24\textwidth}\centering
			\caption{GT}\vspace{-0.2cm}
			\includegraphics[width=.98\textwidth]{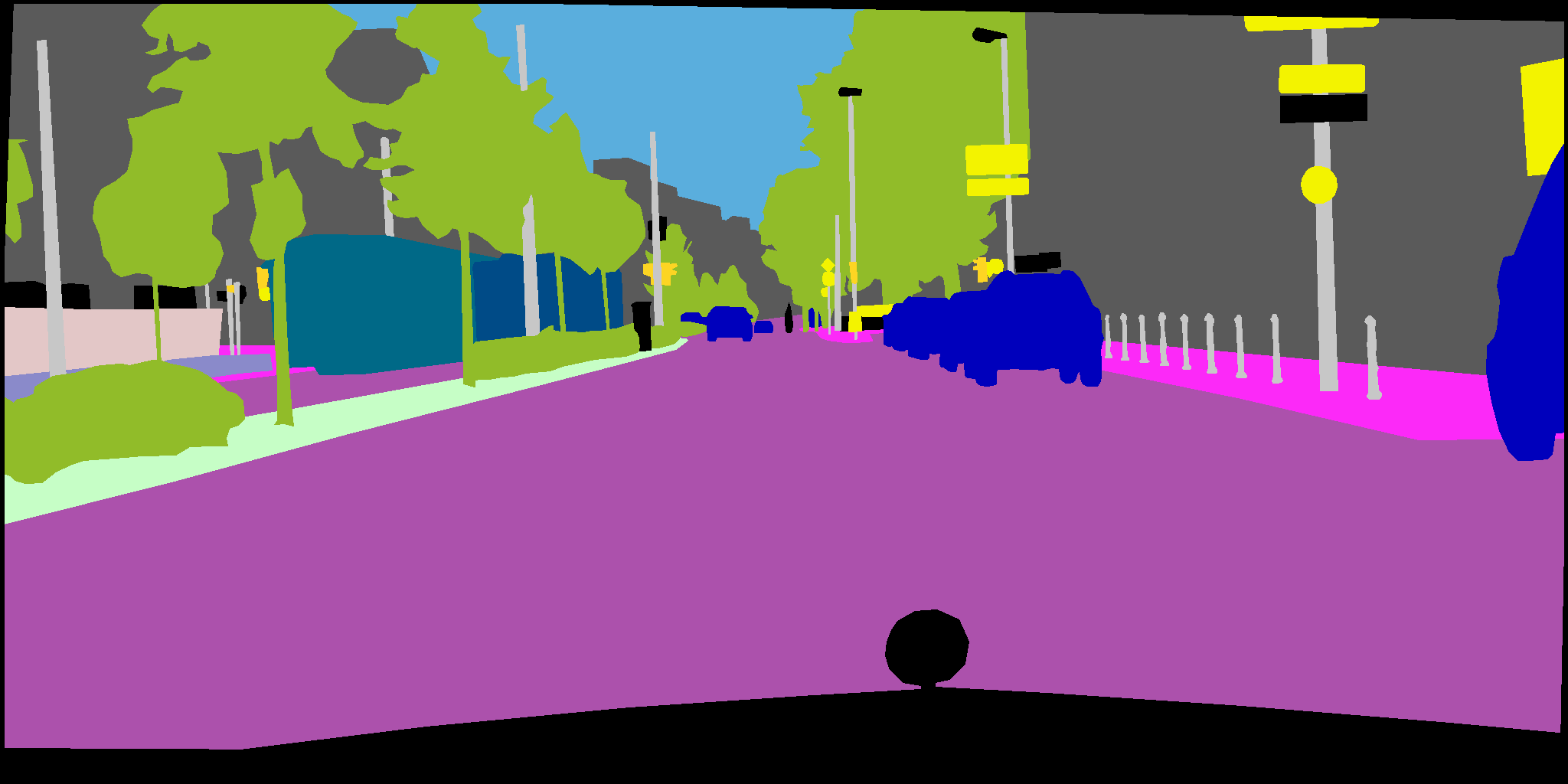}
		\end{subfigure}
		\begin{subfigure}[t]{0.24\textwidth}\centering
			\caption{AdvEnt}\vspace{-0.2cm}
			\includegraphics[width=.98\textwidth]{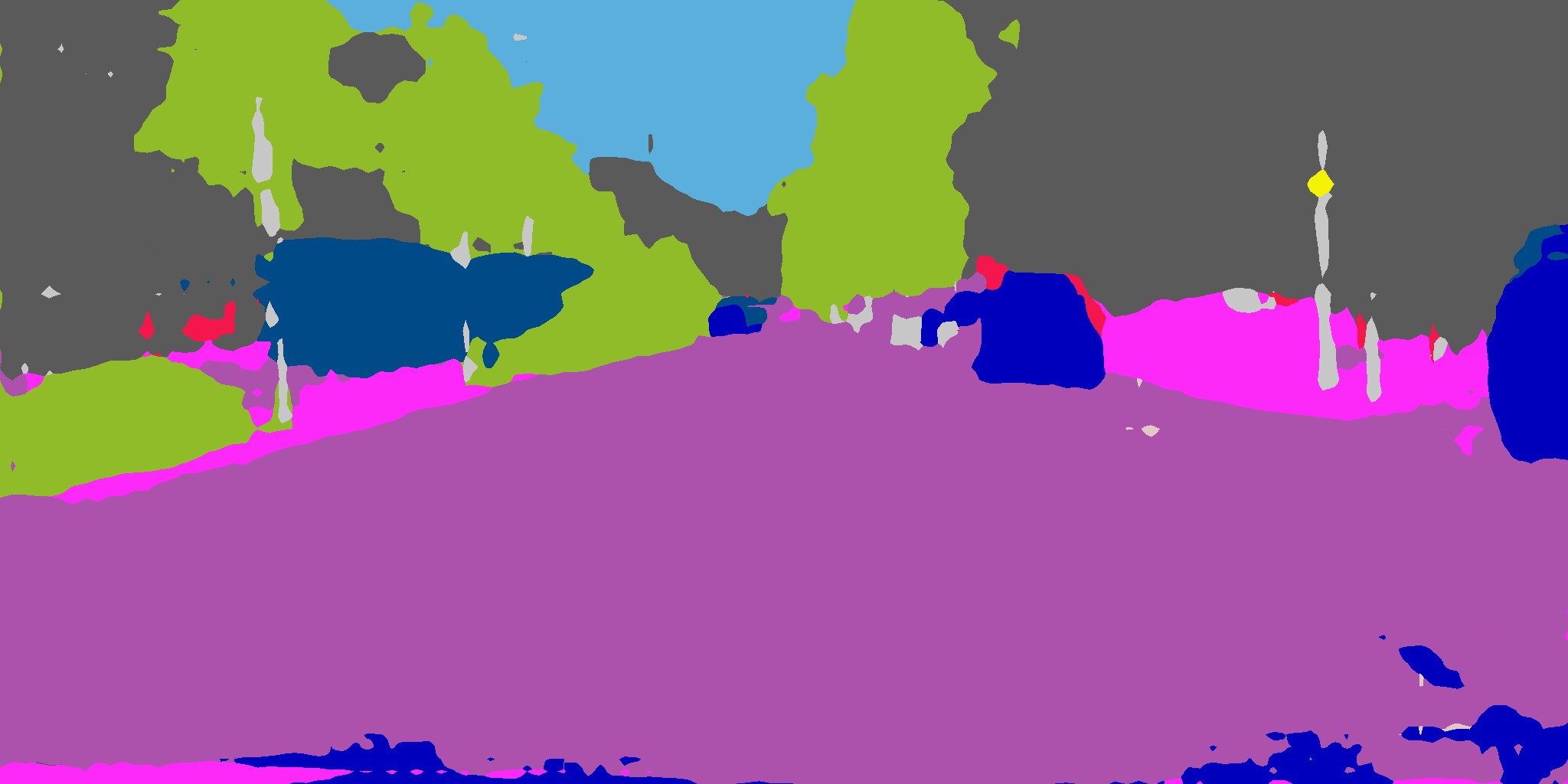}
		\end{subfigure}
		\begin{subfigure}[t]{0.24\textwidth}\centering
			\caption{DADA}\vspace{-0.2cm}
			\includegraphics[width=.98\textwidth]{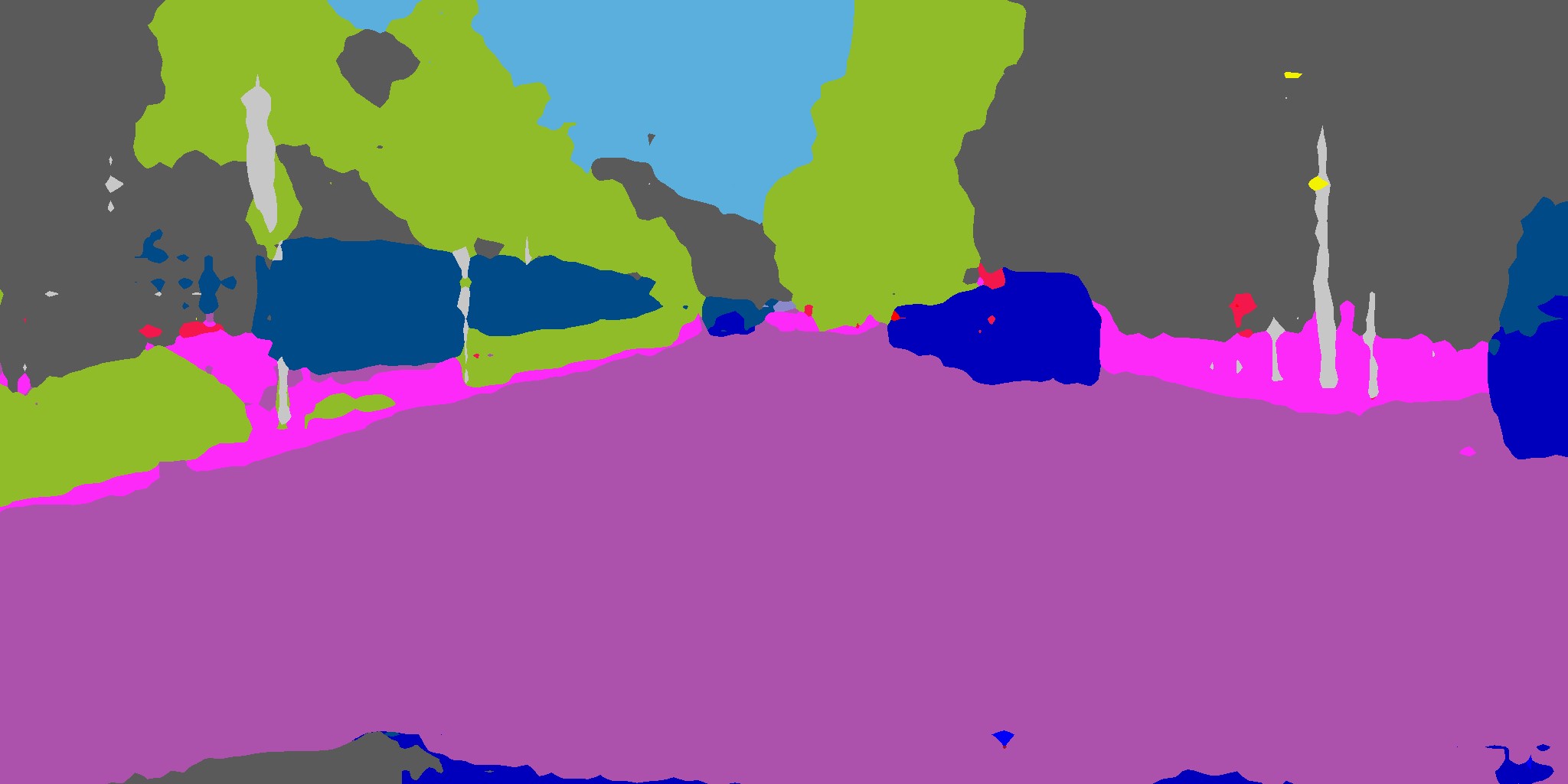}
		\end{subfigure}\\
		\vspace{0.02cm}
		
		\hdashrule[1ex][x]{17cm}{1.5pt}{1.5mm}\vspace{-0.13cm}
		\begin{subfigure}[t]{0.24\textwidth}\centering
			\includegraphics[width=.98\textwidth]{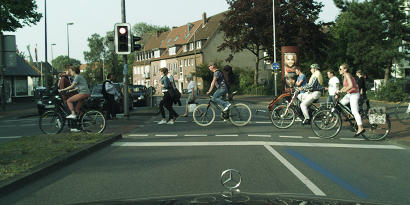}
		\end{subfigure}
		\begin{subfigure}[t]{0.24\textwidth}\centering
			\includegraphics[width=.98\textwidth]{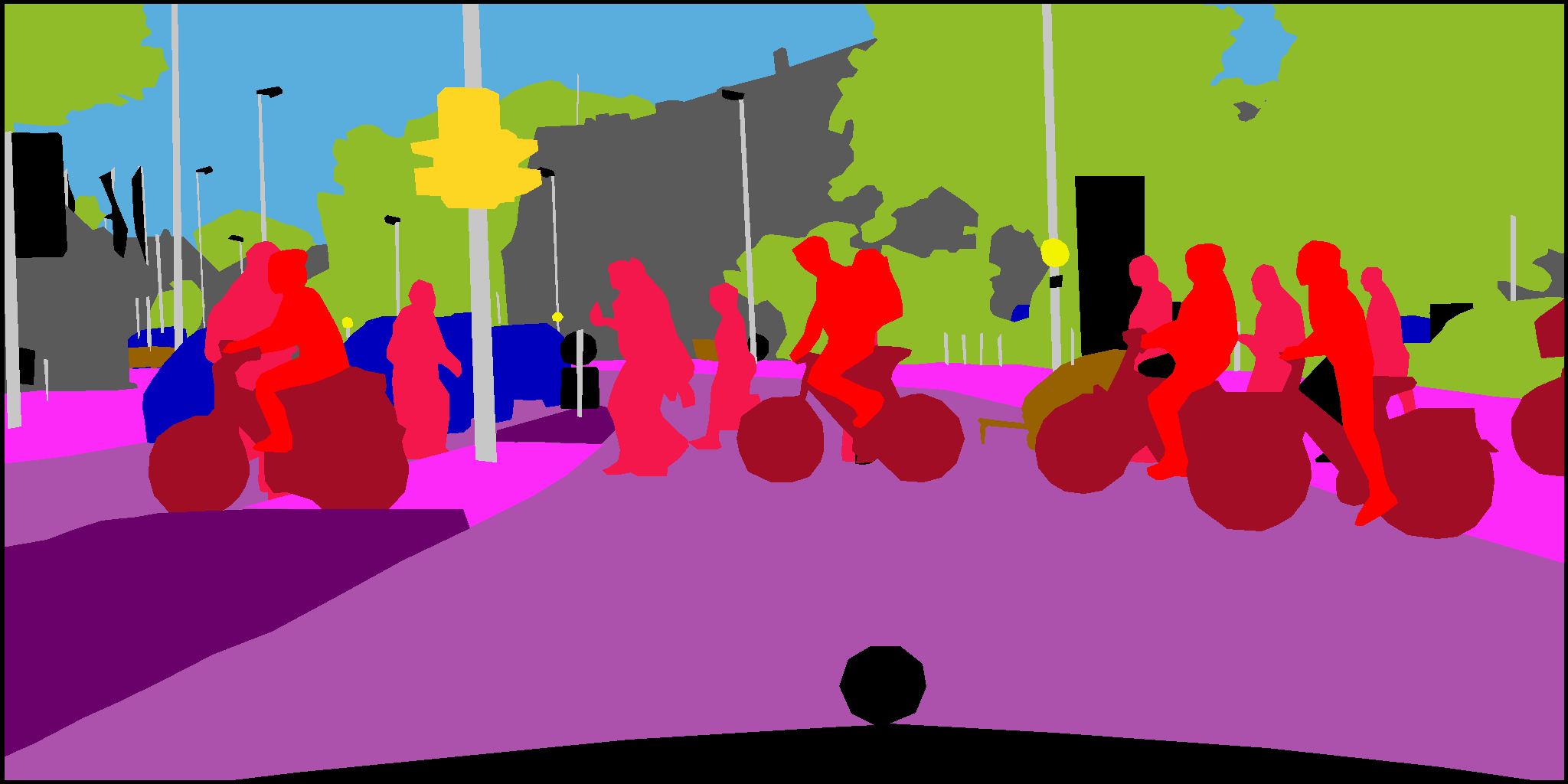}
		\end{subfigure}
		\begin{subfigure}[t]{0.24\textwidth}\centering
			\includegraphics[width=.98\textwidth]{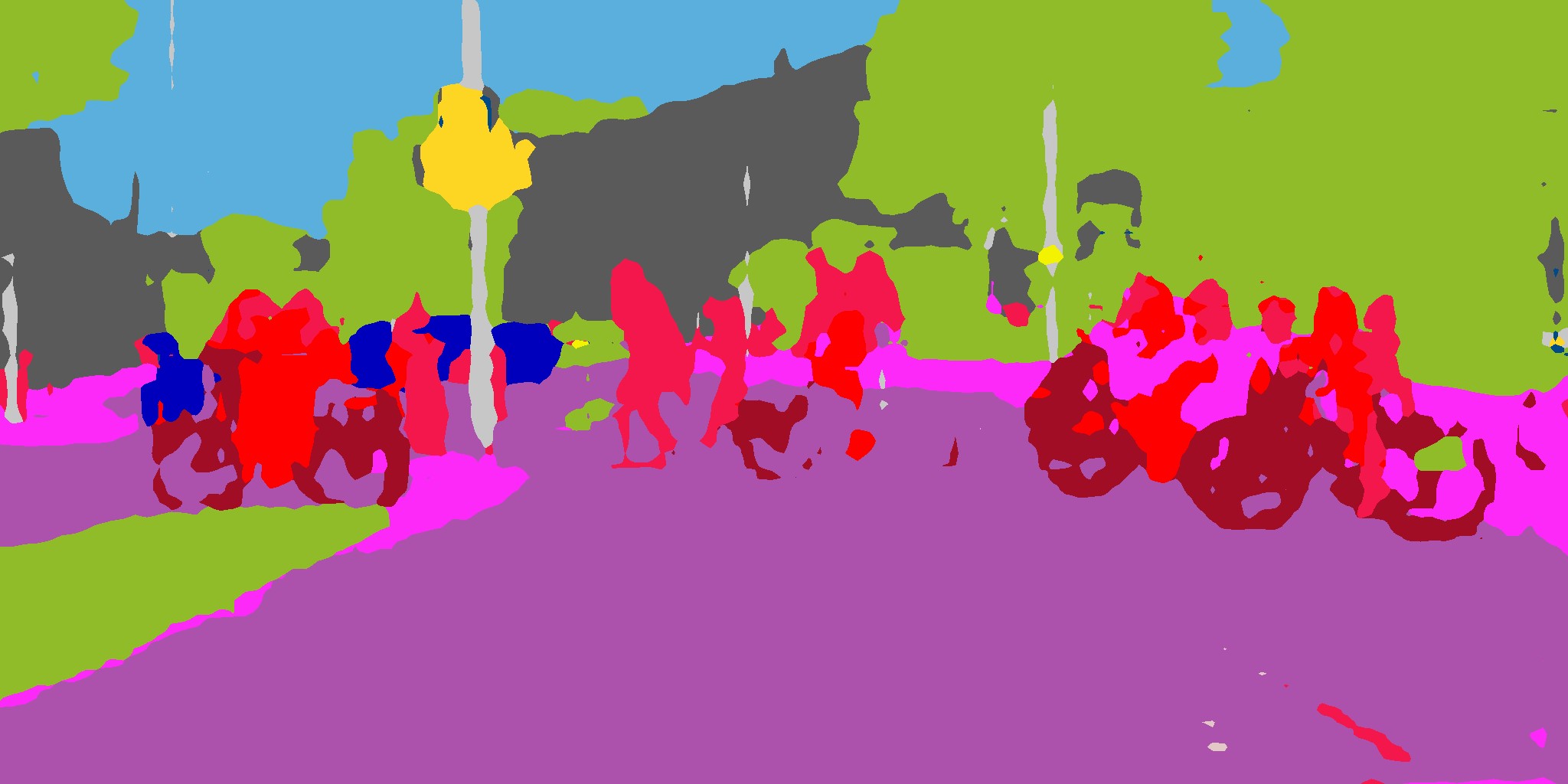}
		\end{subfigure}
		\begin{subfigure}[t]{0.24\textwidth}\centering
			\includegraphics[width=.98\textwidth]{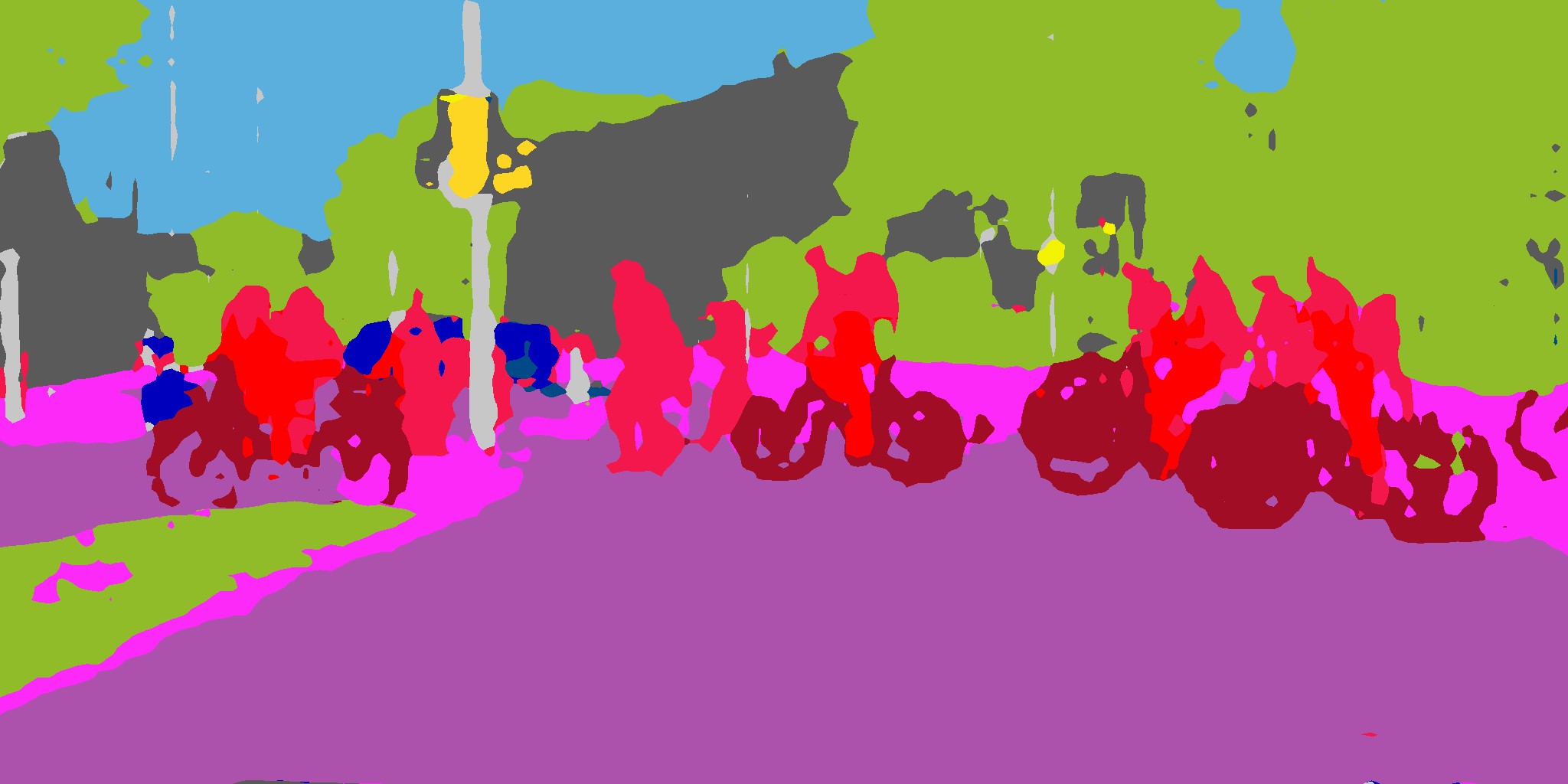}
		\end{subfigure}\\
		\vspace{-0.16cm}
		
		\hdashrule[1ex][x]{17cm}{1.5pt}{1.5mm}\vspace{-0.13cm}
		\begin{subfigure}[t]{0.24\textwidth}\centering
			\includegraphics[width=.98\textwidth]{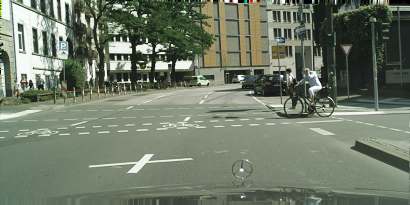}
		\end{subfigure}
		\begin{subfigure}[t]{0.24\textwidth}\centering
			\includegraphics[width=.98\textwidth]{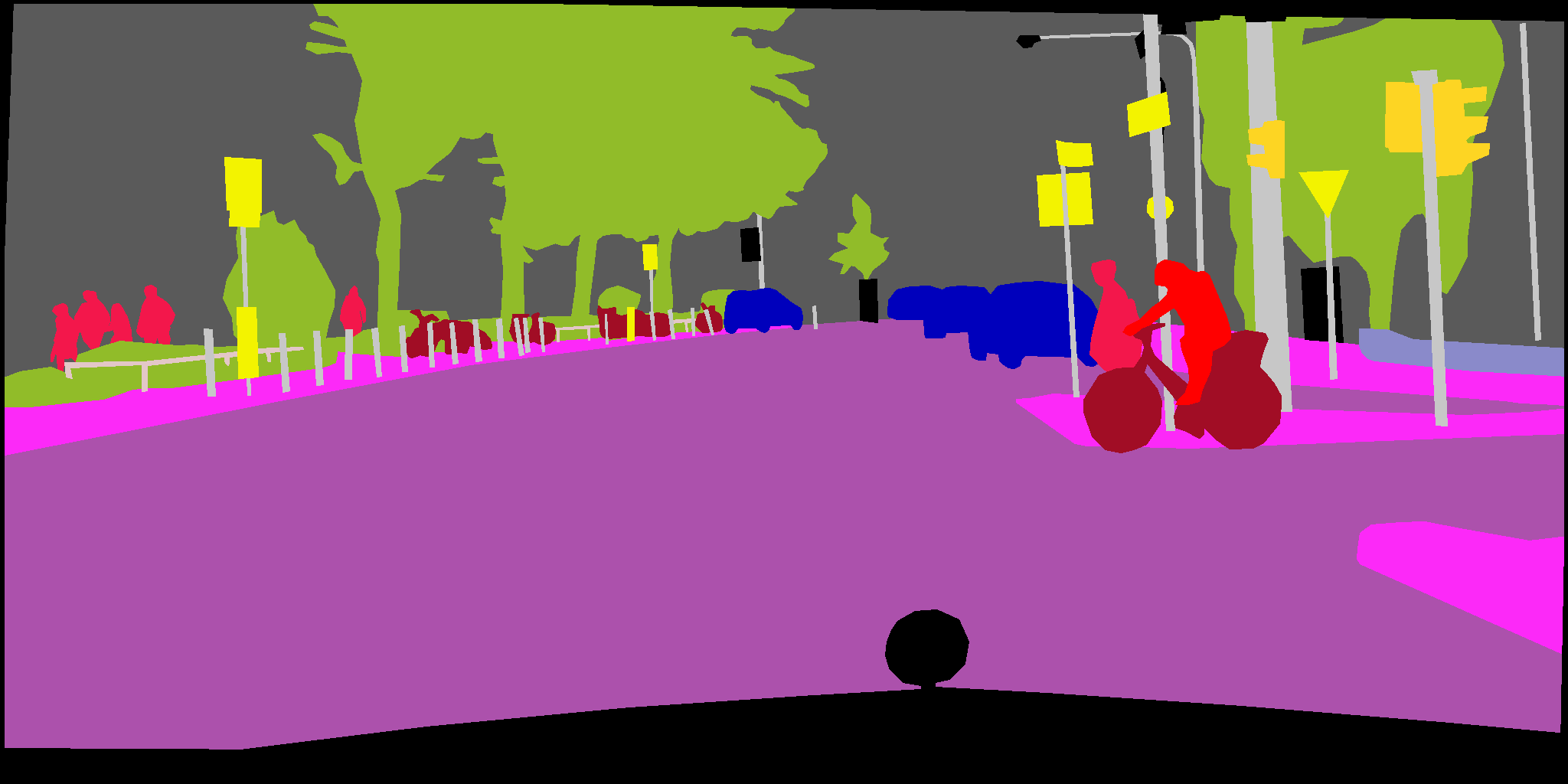}
		\end{subfigure}
		\begin{subfigure}[t]{0.24\textwidth}\centering
			\includegraphics[width=.98\textwidth]{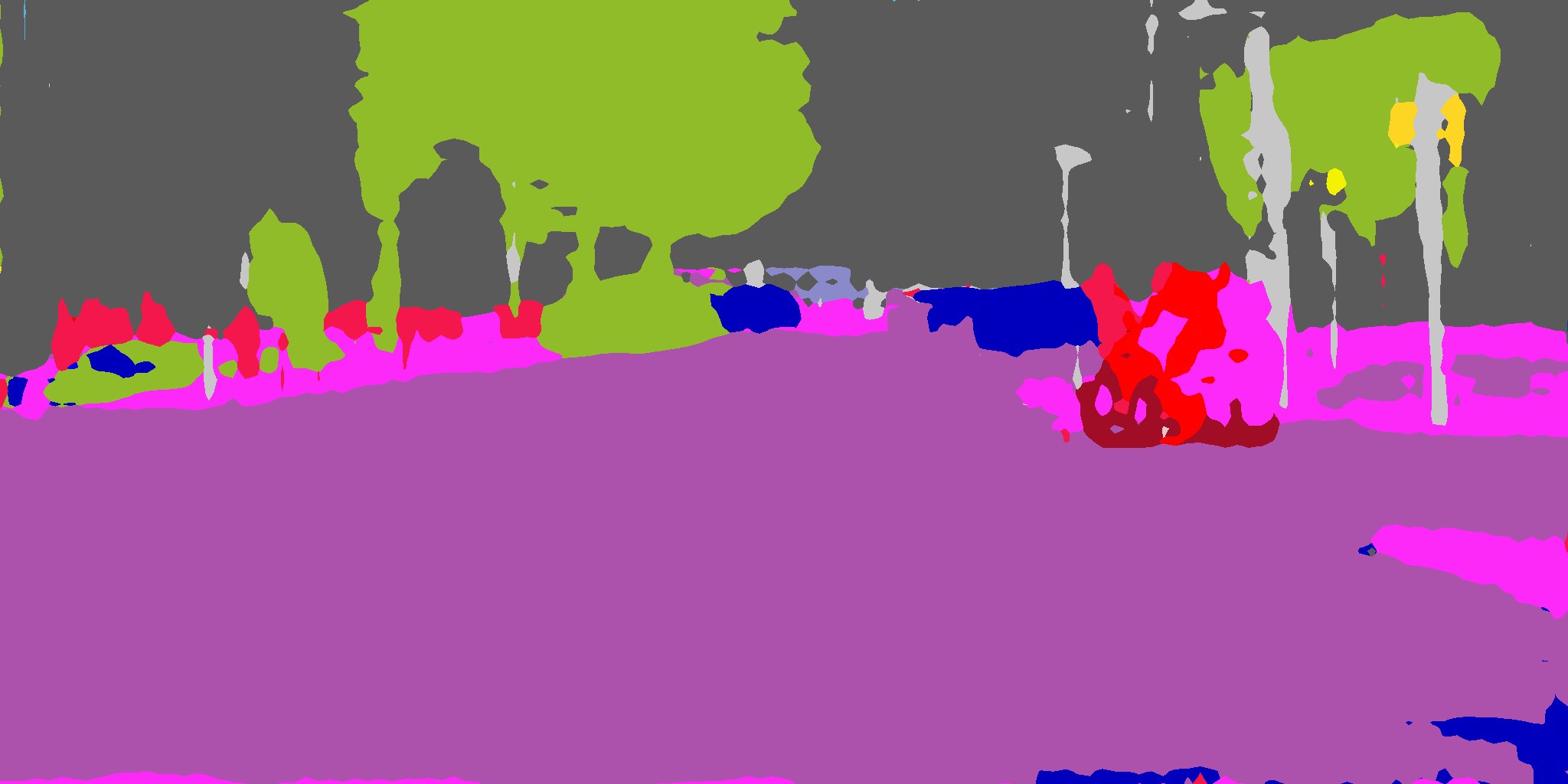}
		\end{subfigure}
		\begin{subfigure}[t]{0.24\textwidth}\centering
			\includegraphics[width=.98\textwidth]{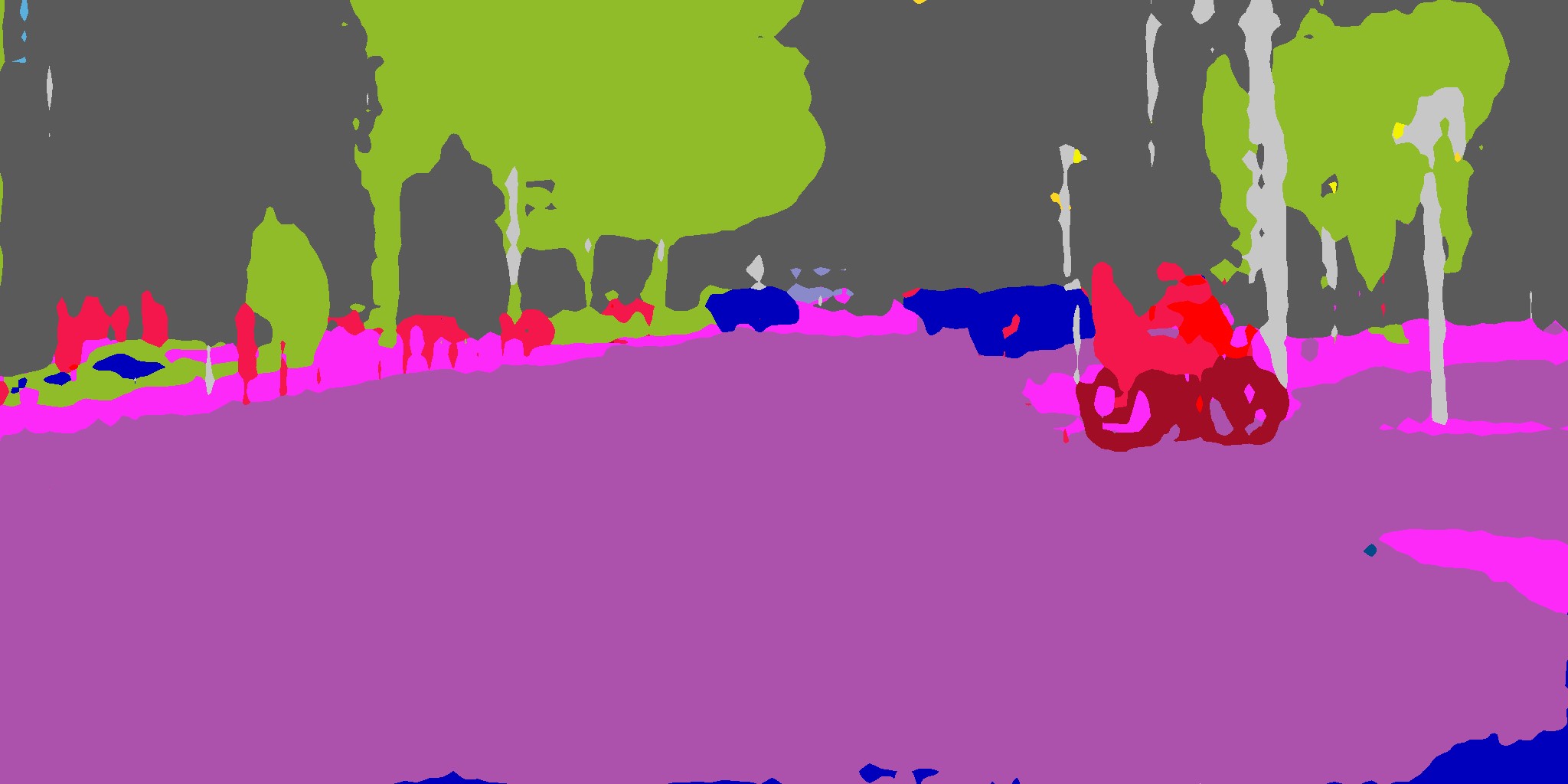}
		\end{subfigure}\\
		\vspace{-0.16cm}
		
		\hdashrule[1ex][x]{17cm}{1.5pt}{1.5mm}\vspace{-0.13cm}
		\begin{subfigure}[t]{0.24\textwidth}\centering
			\includegraphics[width=.98\textwidth]{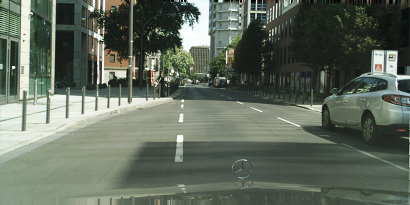}
		\end{subfigure}
		\begin{subfigure}[t]{0.24\textwidth}\centering
			\includegraphics[width=.98\textwidth]{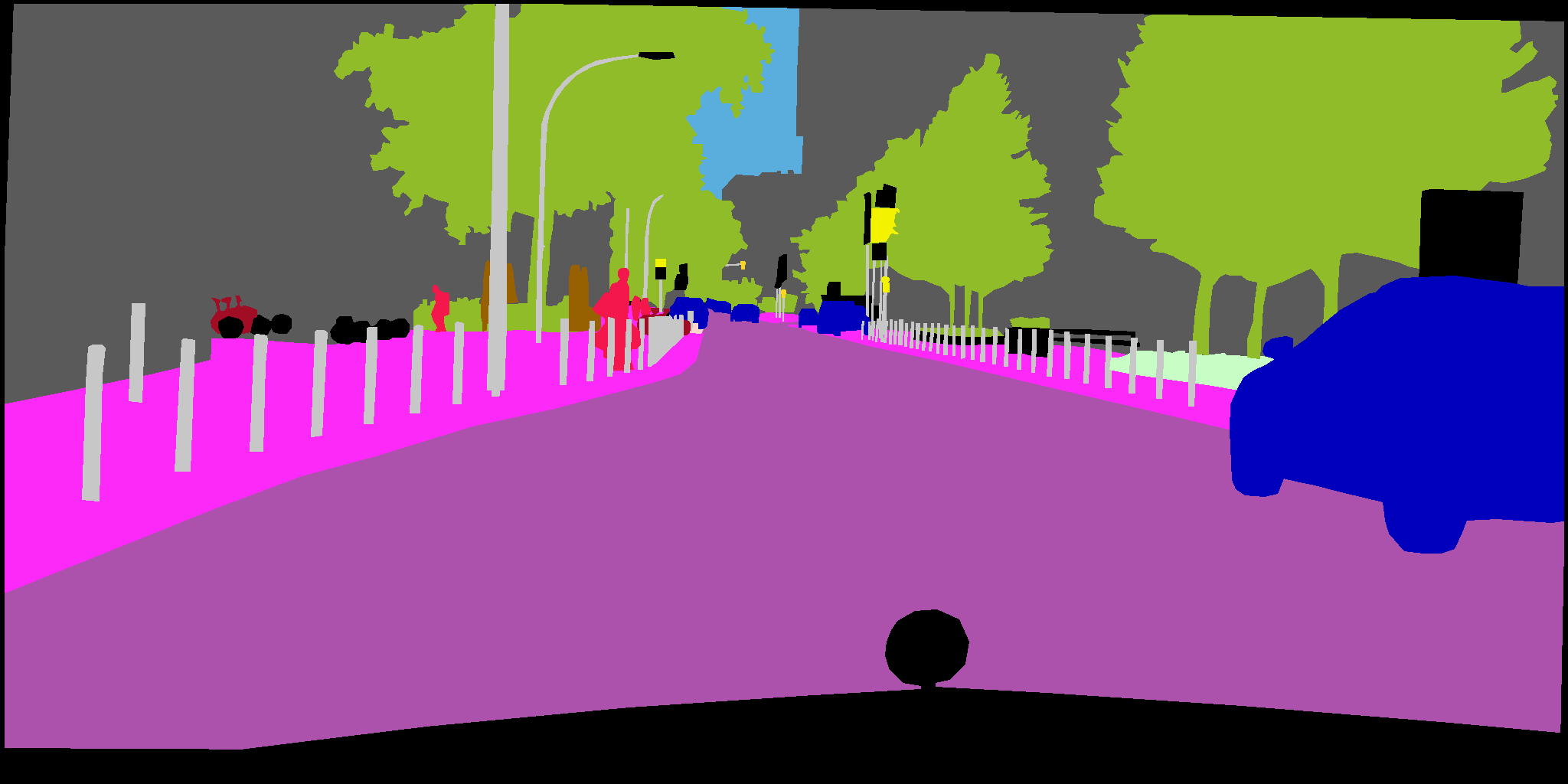}
		\end{subfigure}
		\begin{subfigure}[t]{0.24\textwidth}\centering
			\includegraphics[width=.98\textwidth]{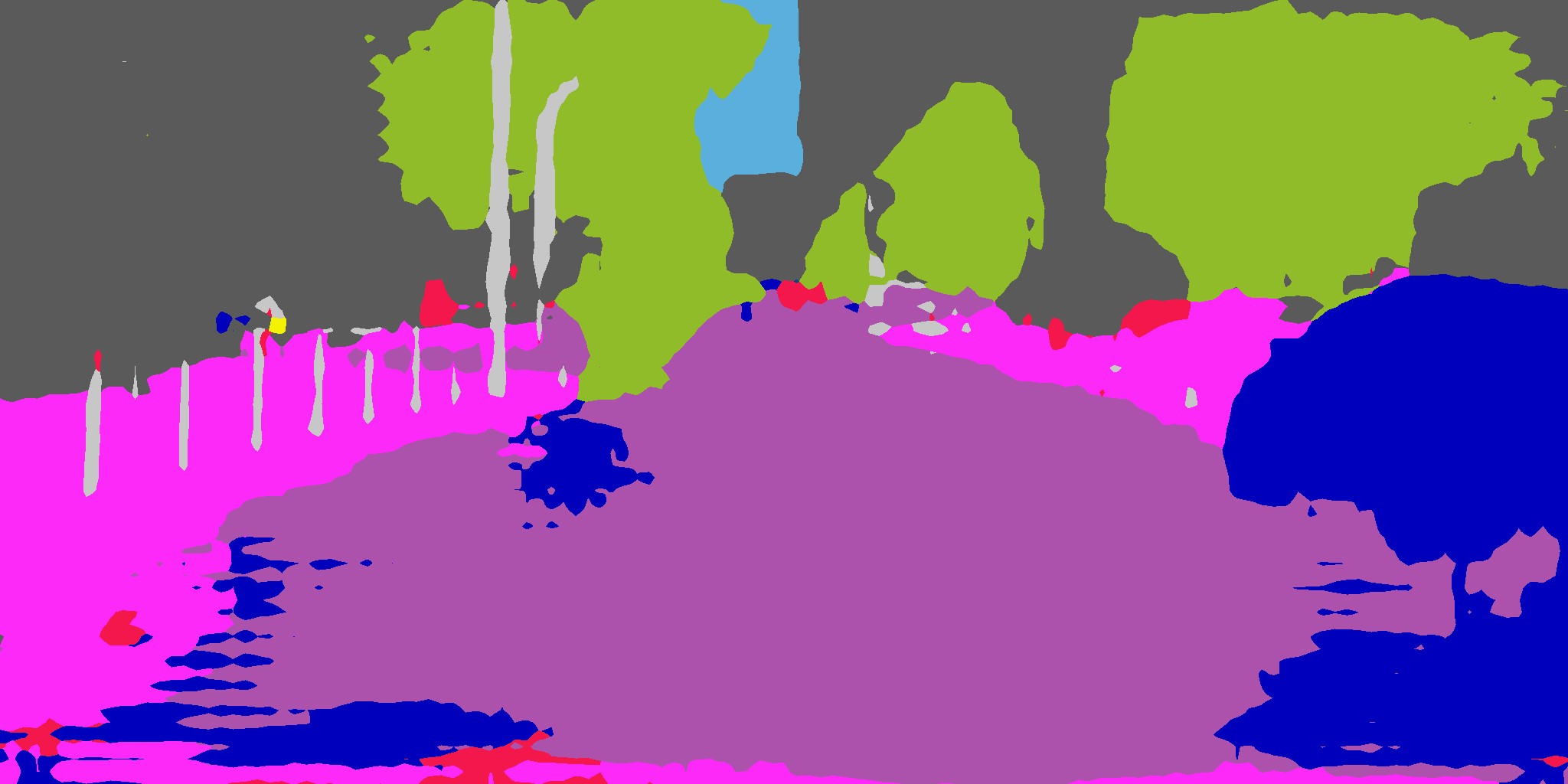}
		\end{subfigure}
		\begin{subfigure}[t]{0.24\textwidth}\centering
			\includegraphics[width=.98\textwidth]{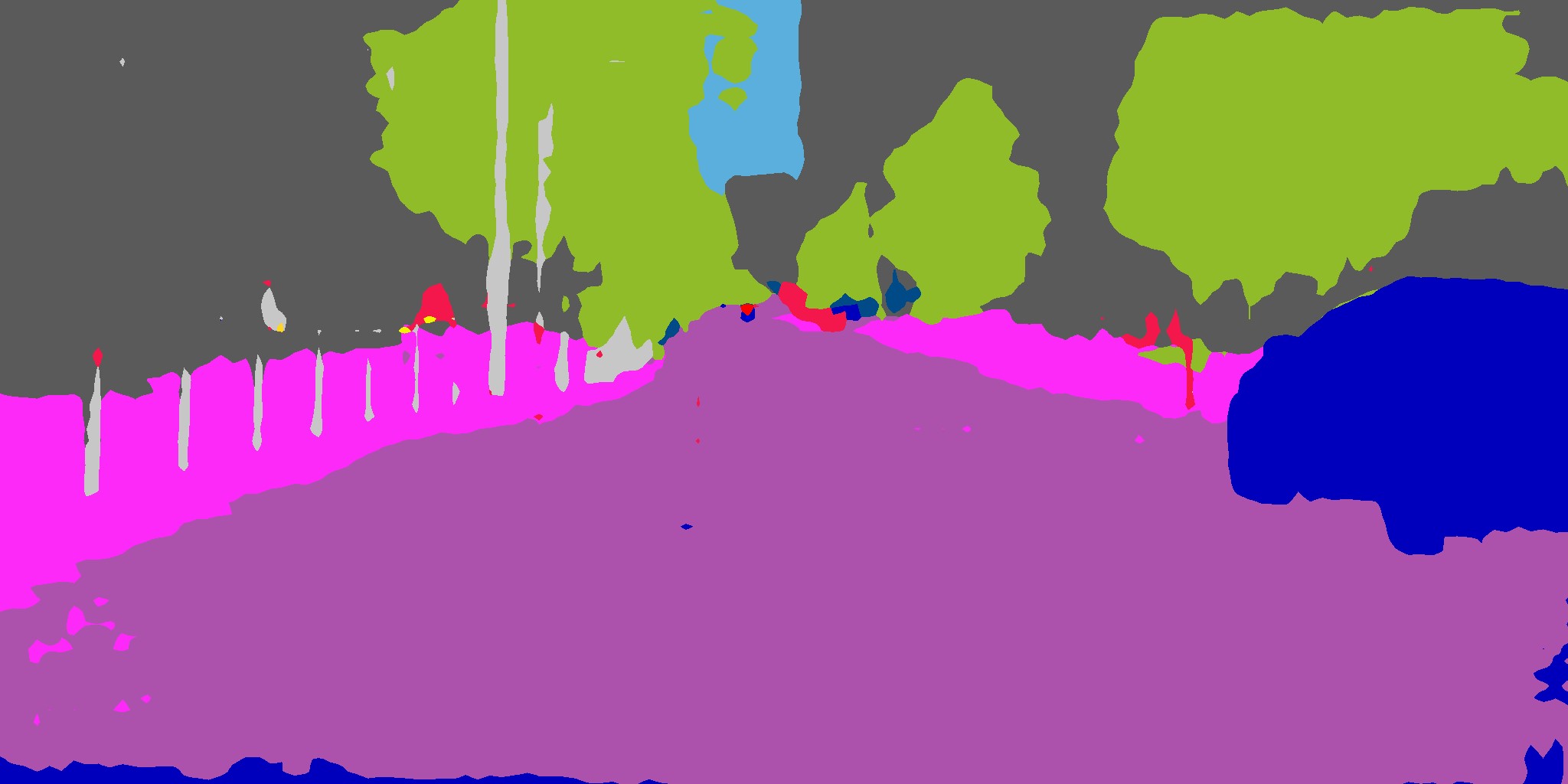}
		\end{subfigure}
		
		\hdashrule[1ex][x]{17cm}{1.5pt}{1.5mm}\vspace{-0.13cm}
		\begin{subfigure}[t]{0.24\textwidth}\centering
			\includegraphics[width=.98\textwidth]{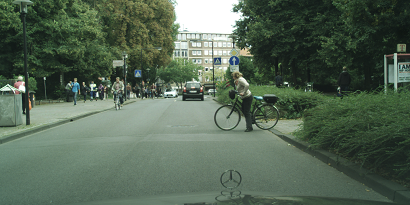}
		\end{subfigure}
		\begin{subfigure}[t]{0.24\textwidth}\centering
			\includegraphics[width=.98\textwidth]{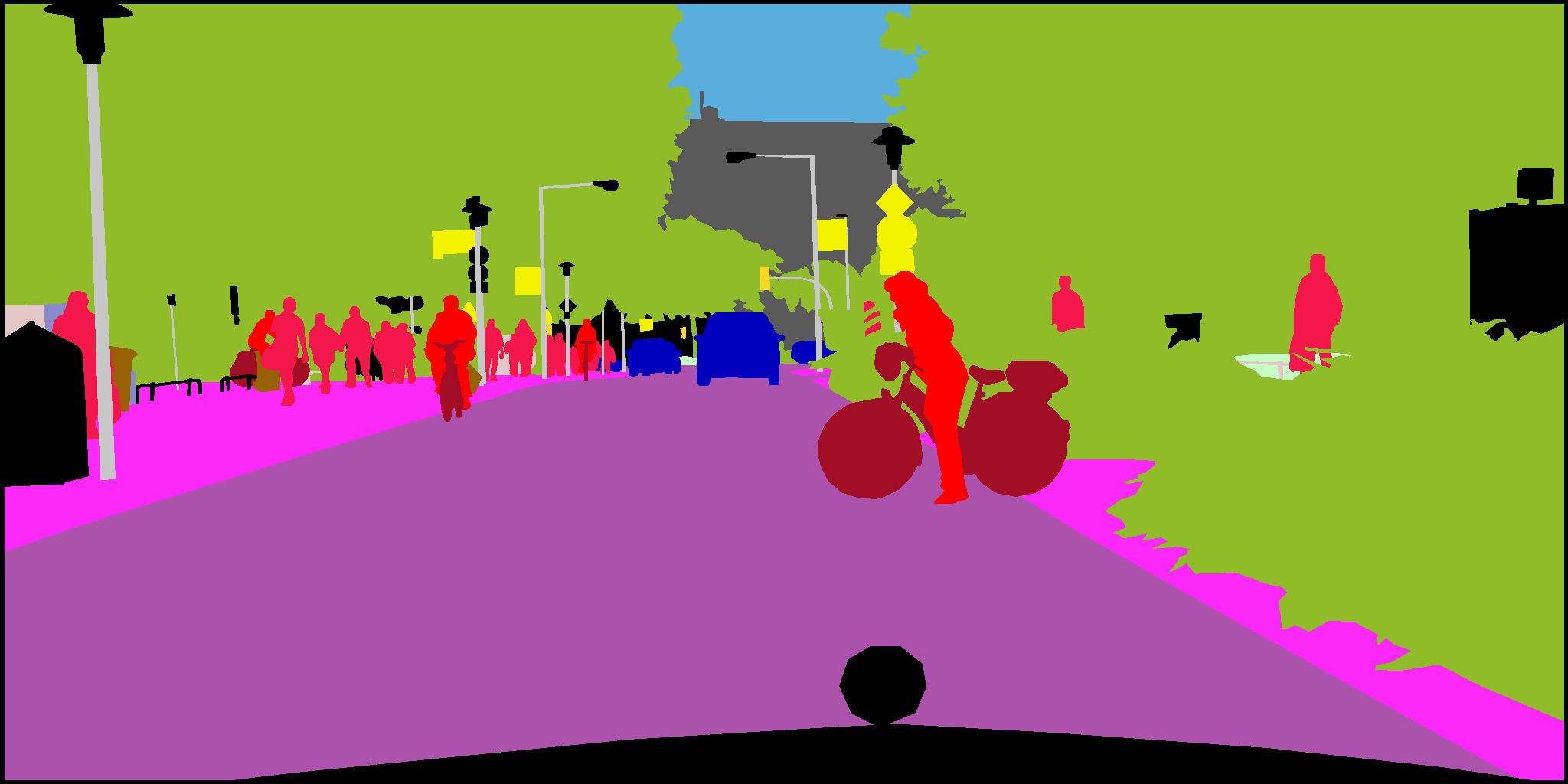}
		\end{subfigure}
		\begin{subfigure}[t]{0.24\textwidth}\centering
			\includegraphics[width=.98\textwidth]{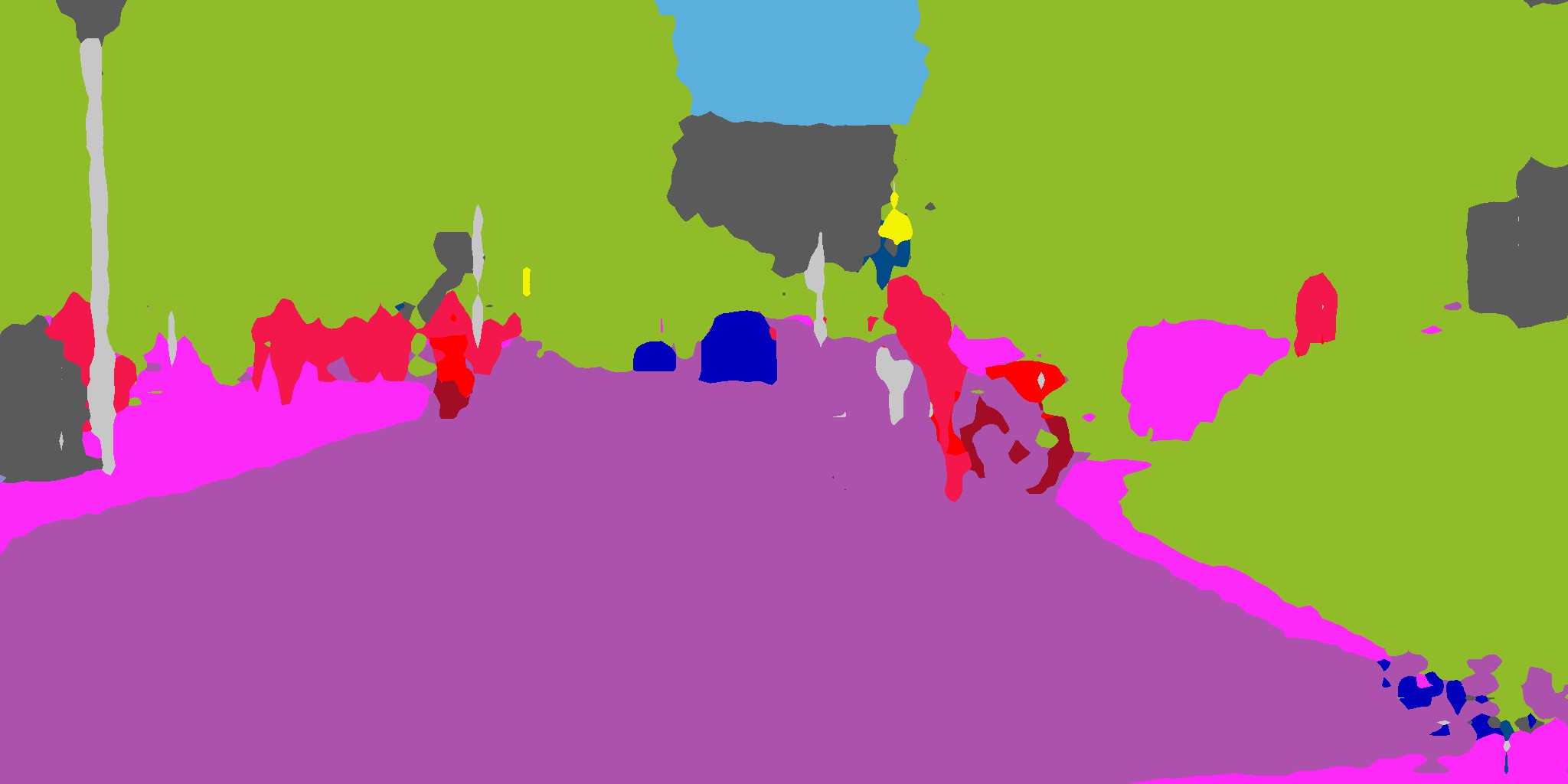}
		\end{subfigure}
		\begin{subfigure}[t]{0.24\textwidth}\centering
			\includegraphics[width=.98\textwidth]{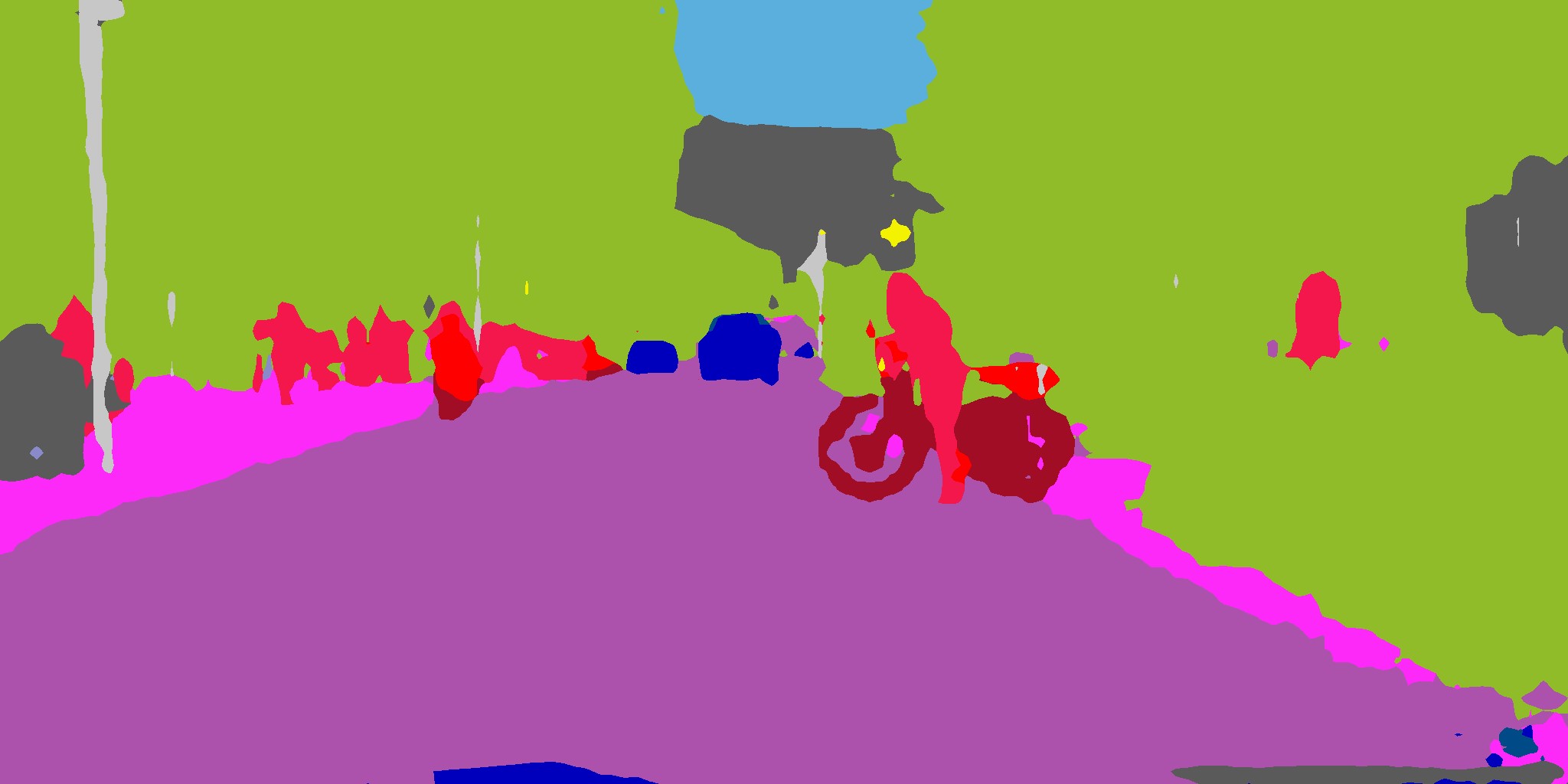}
		\end{subfigure}
		
		\hdashrule[1ex][x]{17cm}{1.5pt}{1.5mm}\vspace{-0.13cm}
		\begin{subfigure}[t]{0.24\textwidth}\centering
			\includegraphics[width=.98\textwidth]{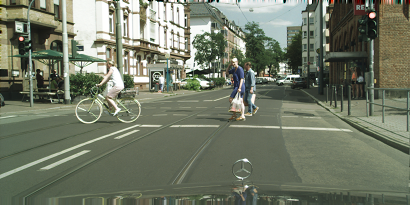}
		\end{subfigure}
		\begin{subfigure}[t]{0.24\textwidth}\centering
			\includegraphics[width=.98\textwidth]{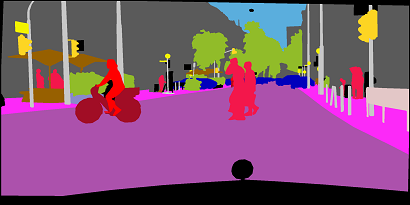}
		\end{subfigure}
		\begin{subfigure}[t]{0.24\textwidth}\centering
			\includegraphics[width=.98\textwidth]{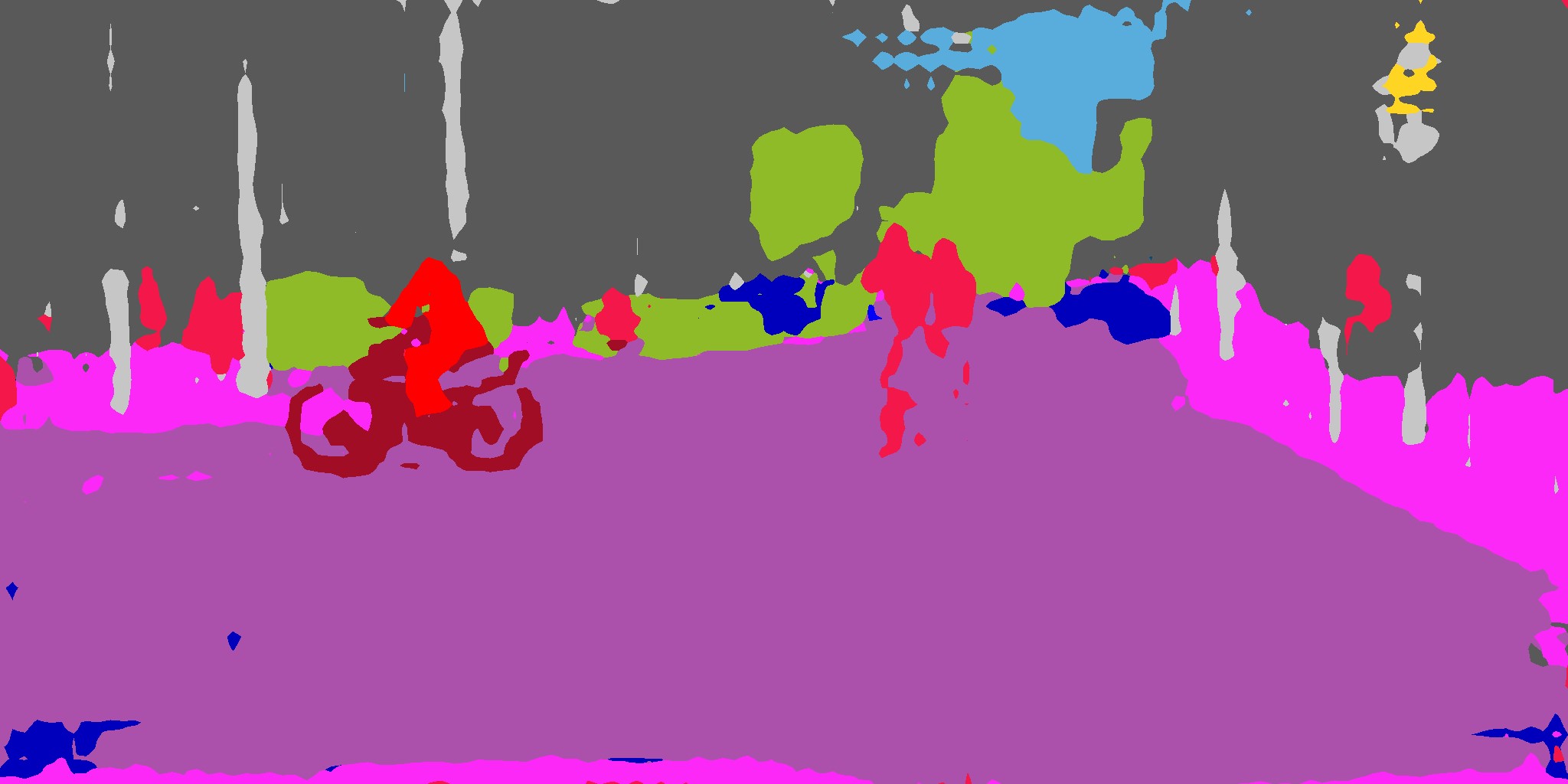}
		\end{subfigure}
		\begin{subfigure}[t]{0.24\textwidth}\centering
			\includegraphics[width=.98\textwidth]{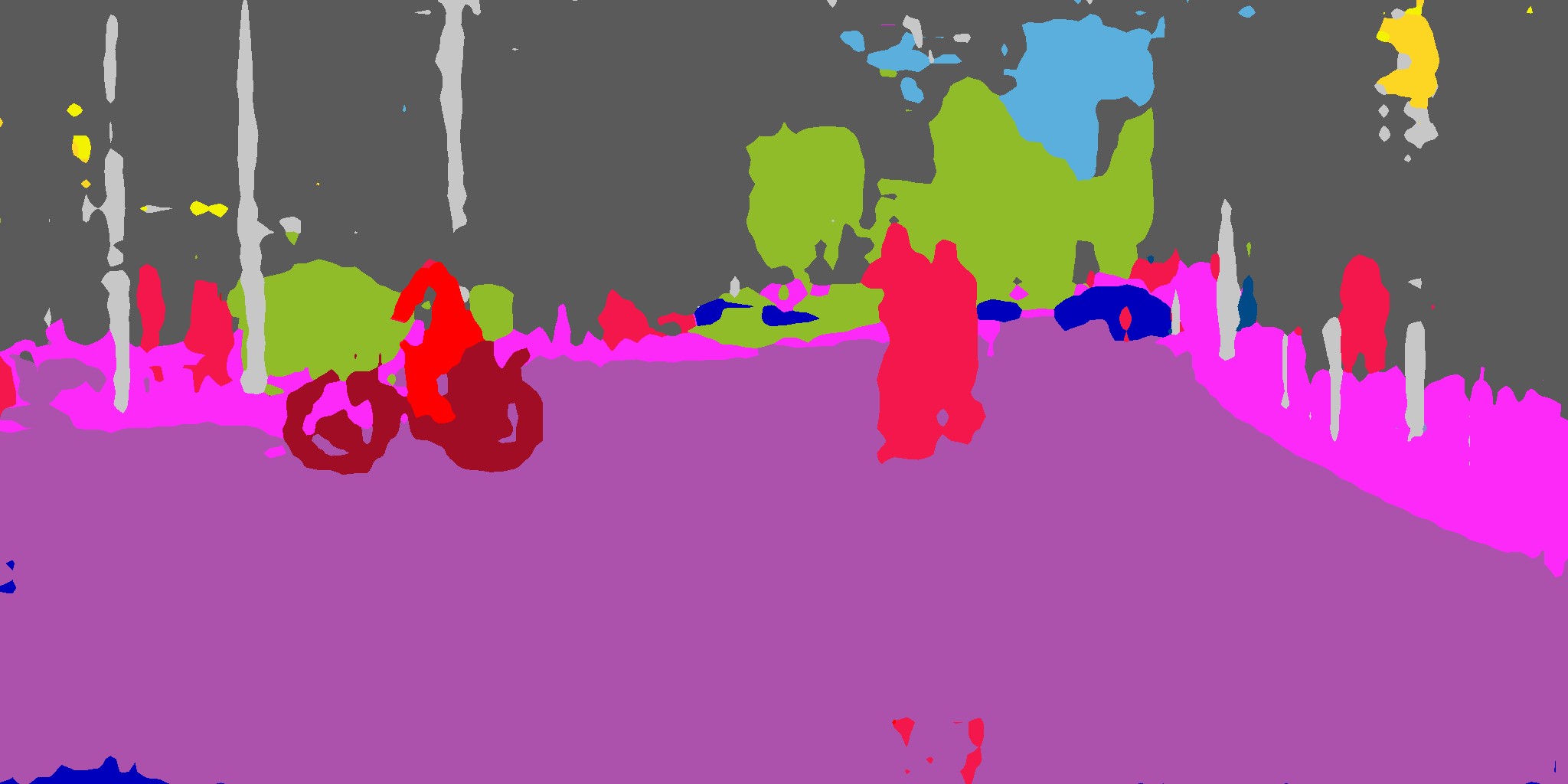}
		\end{subfigure}	
		
		\hdashrule[1ex][x]{17cm}{1.5pt}{1.5mm}\vspace{-0.13cm}
		\begin{subfigure}[t]{0.24\textwidth}\centering
			\includegraphics[width=.98\textwidth]{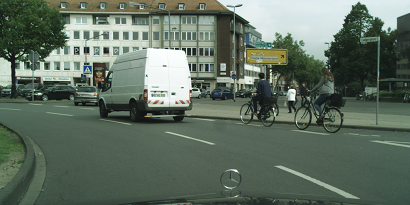}
		\end{subfigure}
		\begin{subfigure}[t]{0.24\textwidth}\centering
			\includegraphics[width=.98\textwidth]{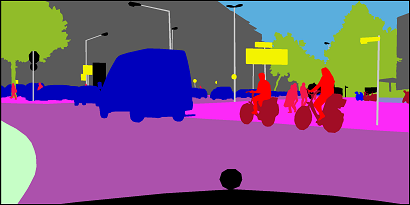}
		\end{subfigure}
		\begin{subfigure}[t]{0.24\textwidth}\centering
			\includegraphics[width=.98\textwidth]{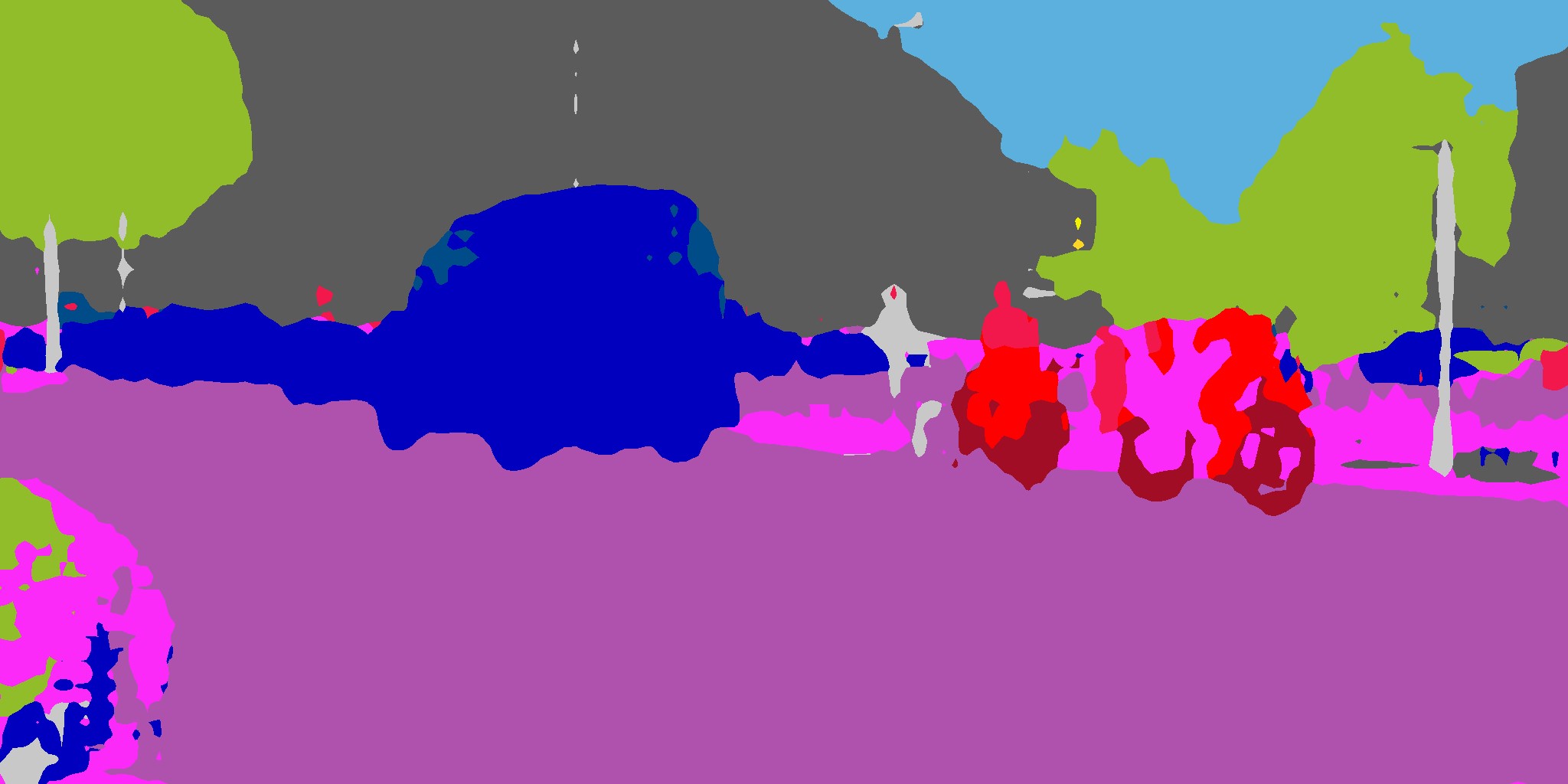}
		\end{subfigure}
		\begin{subfigure}[t]{0.24\textwidth}\centering
			\includegraphics[width=.98\textwidth]{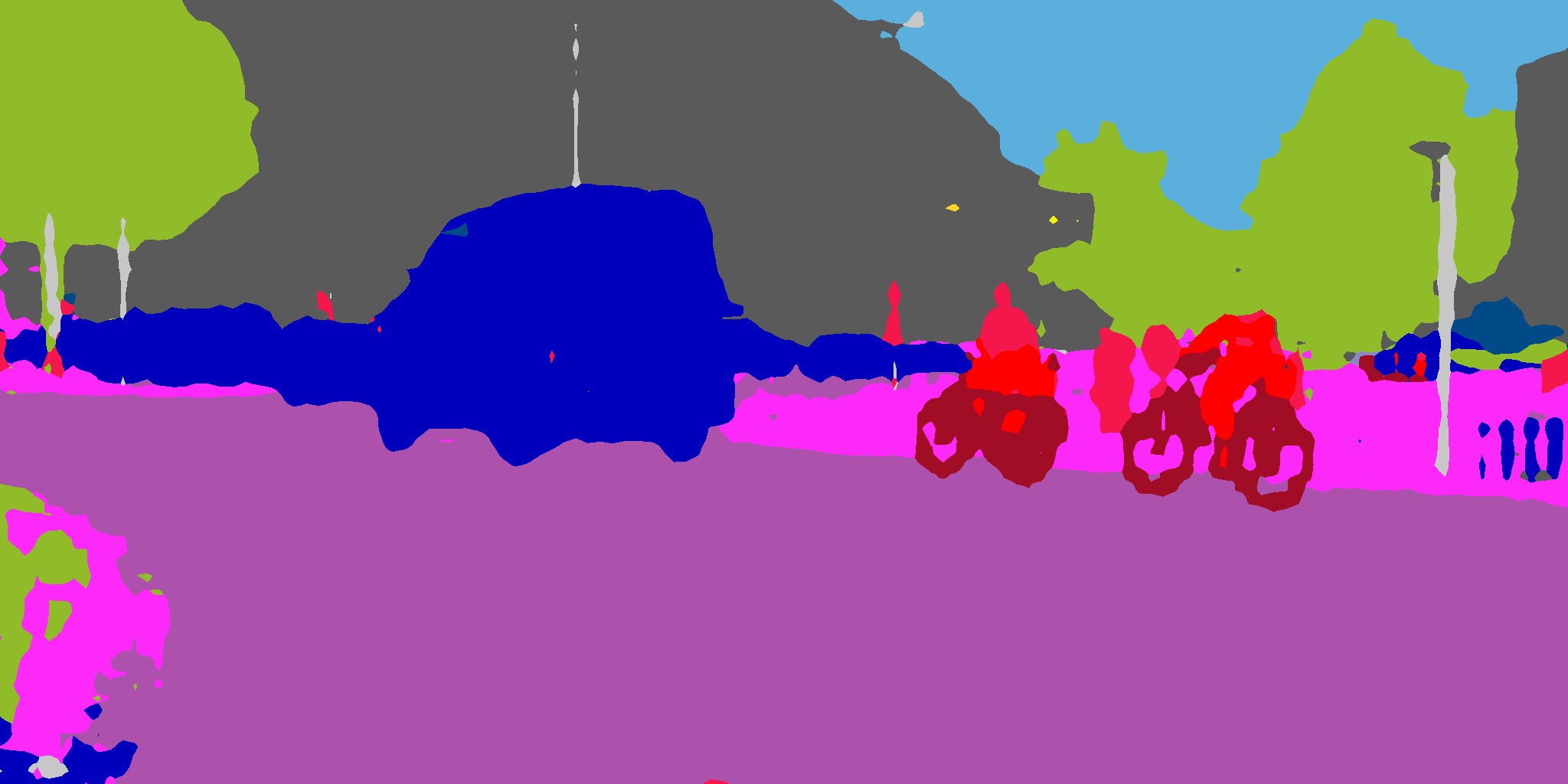}
		\end{subfigure}	
		
		\hdashrule[1ex][x]{17cm}{1.5pt}{1.5mm}\vspace{-0.13cm}
		\begin{subfigure}[t]{0.24\textwidth}\centering
			\includegraphics[width=.98\textwidth]{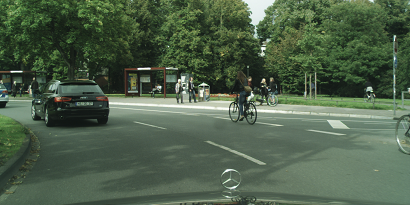}
		\end{subfigure}
		\begin{subfigure}[t]{0.24\textwidth}\centering
			\includegraphics[width=.98\textwidth]{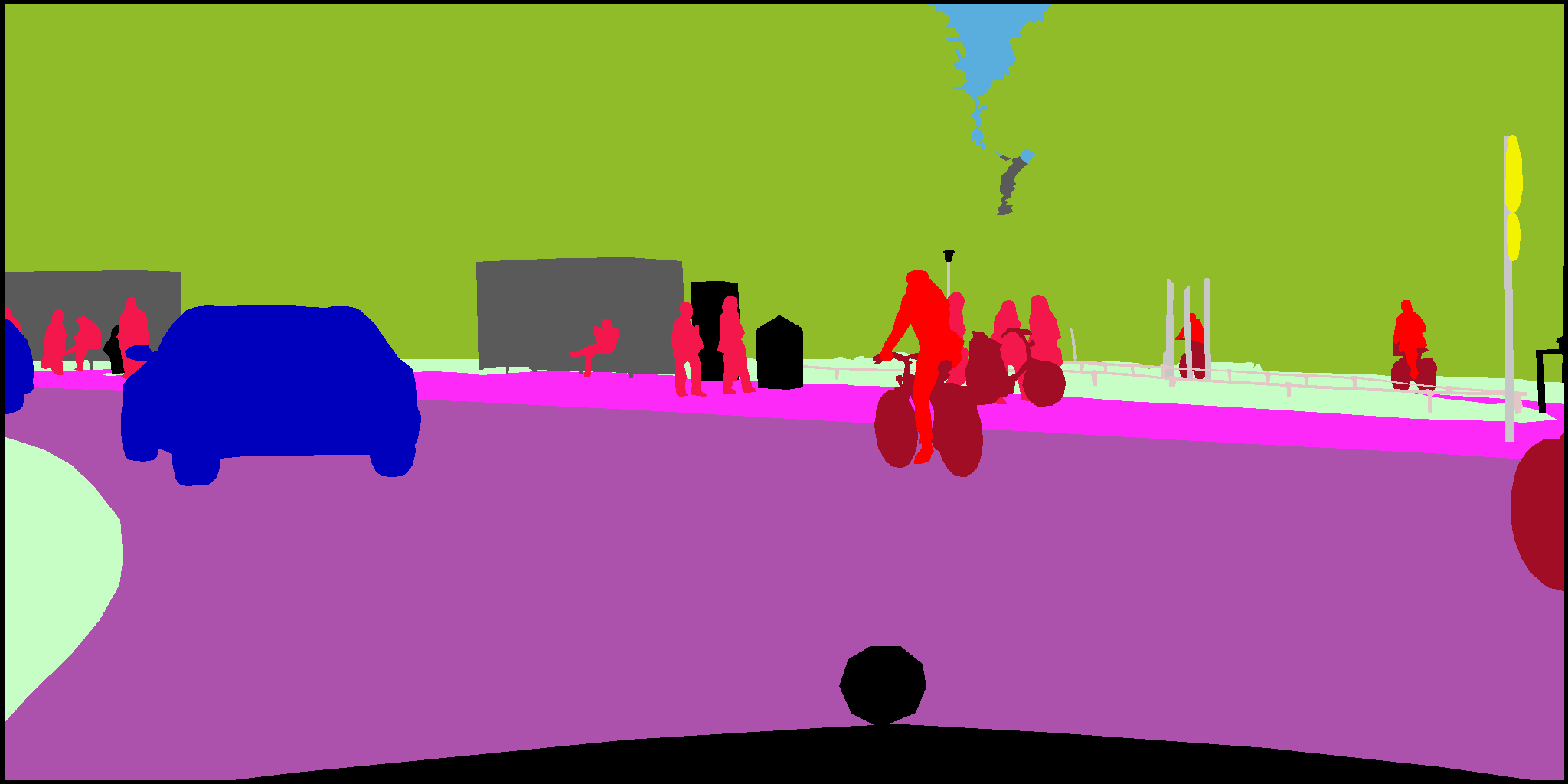}
		\end{subfigure}
		\begin{subfigure}[t]{0.24\textwidth}\centering
			\includegraphics[width=.98\textwidth]{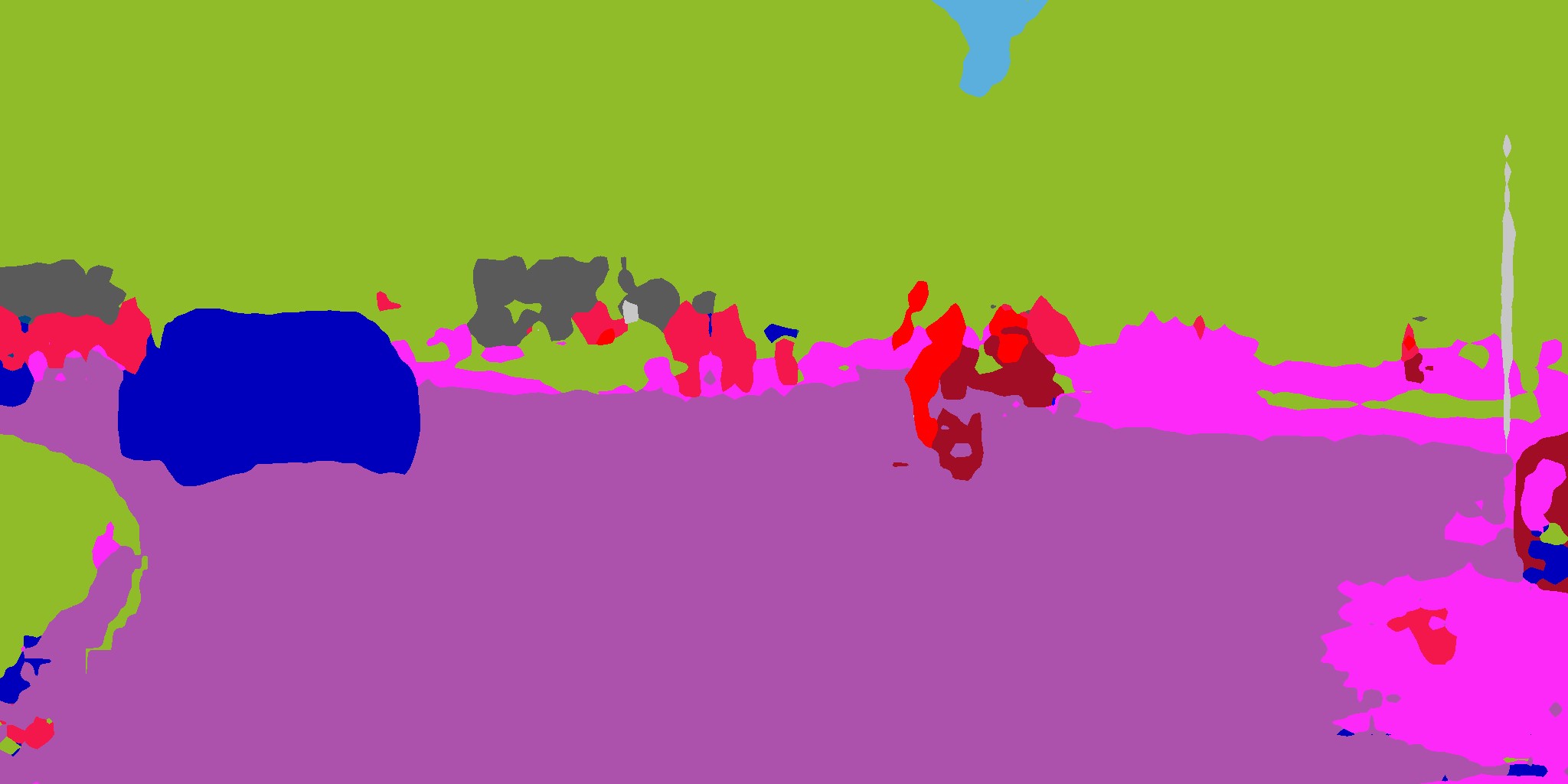}
		\end{subfigure}
		\begin{subfigure}[t]{0.24\textwidth}\centering
			\includegraphics[width=.98\textwidth]{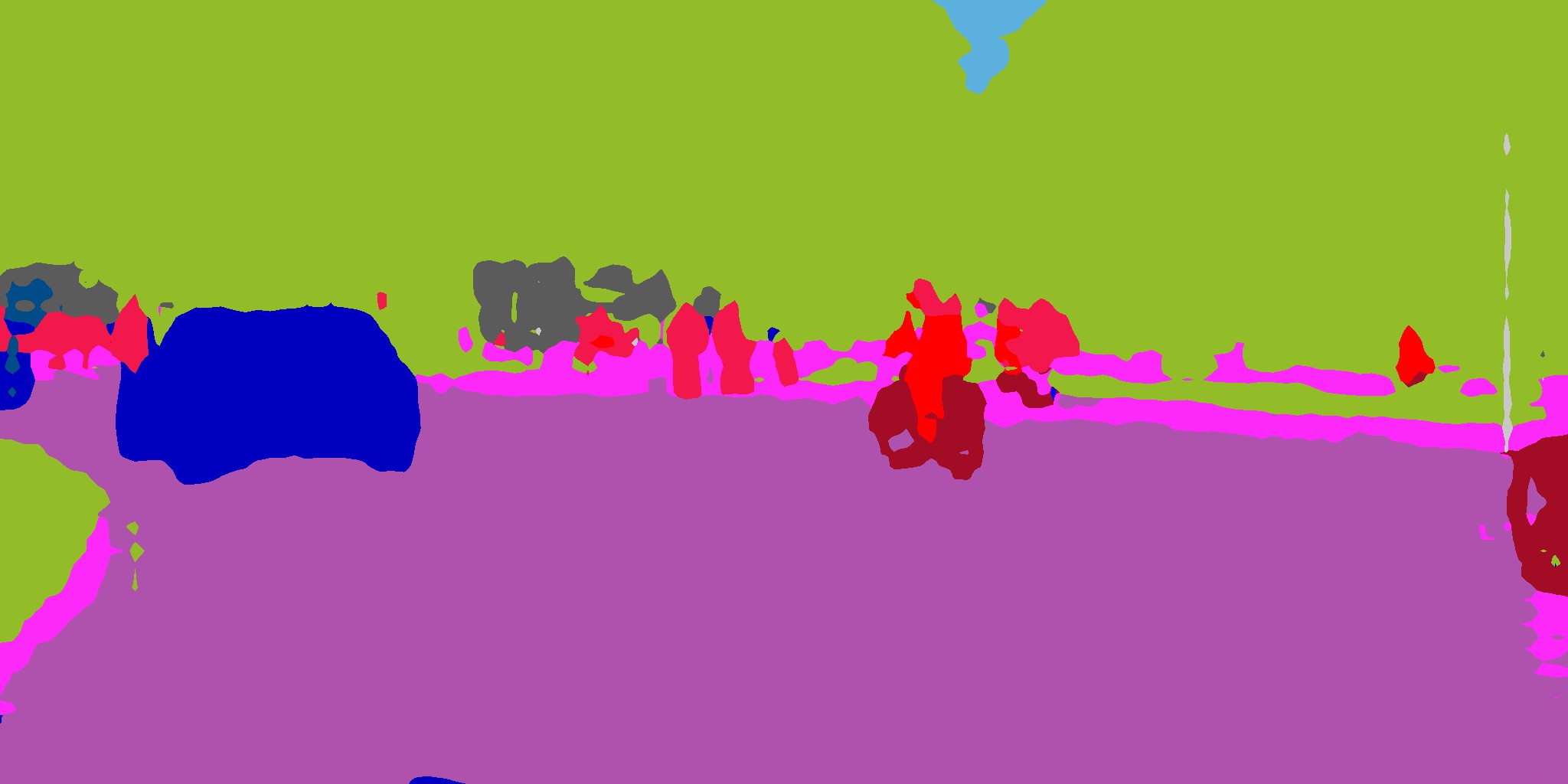}
		\end{subfigure}
		
		\hdashrule[1ex][x]{17cm}{1.5pt}{1.5mm}\vspace{-0.13cm}
		\begin{subfigure}[t]{0.24\textwidth}\centering
			\includegraphics[width=.98\textwidth]{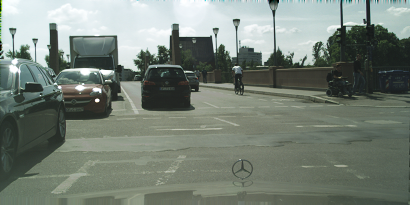}
		\end{subfigure}
		\begin{subfigure}[t]{0.24\textwidth}\centering
			\includegraphics[width=.98\textwidth]{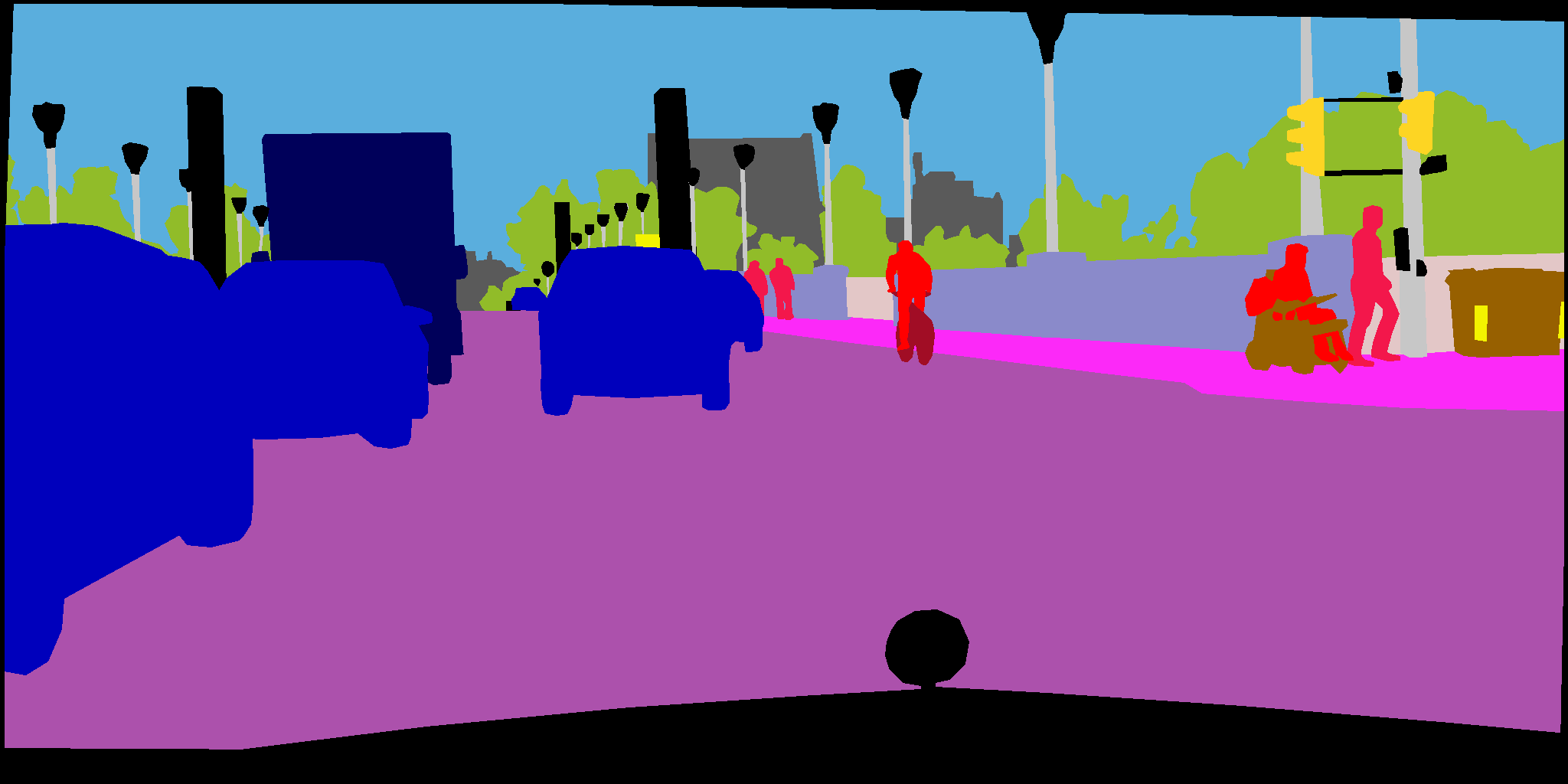}
		\end{subfigure}
		\begin{subfigure}[t]{0.24\textwidth}\centering
			\includegraphics[width=.98\textwidth]{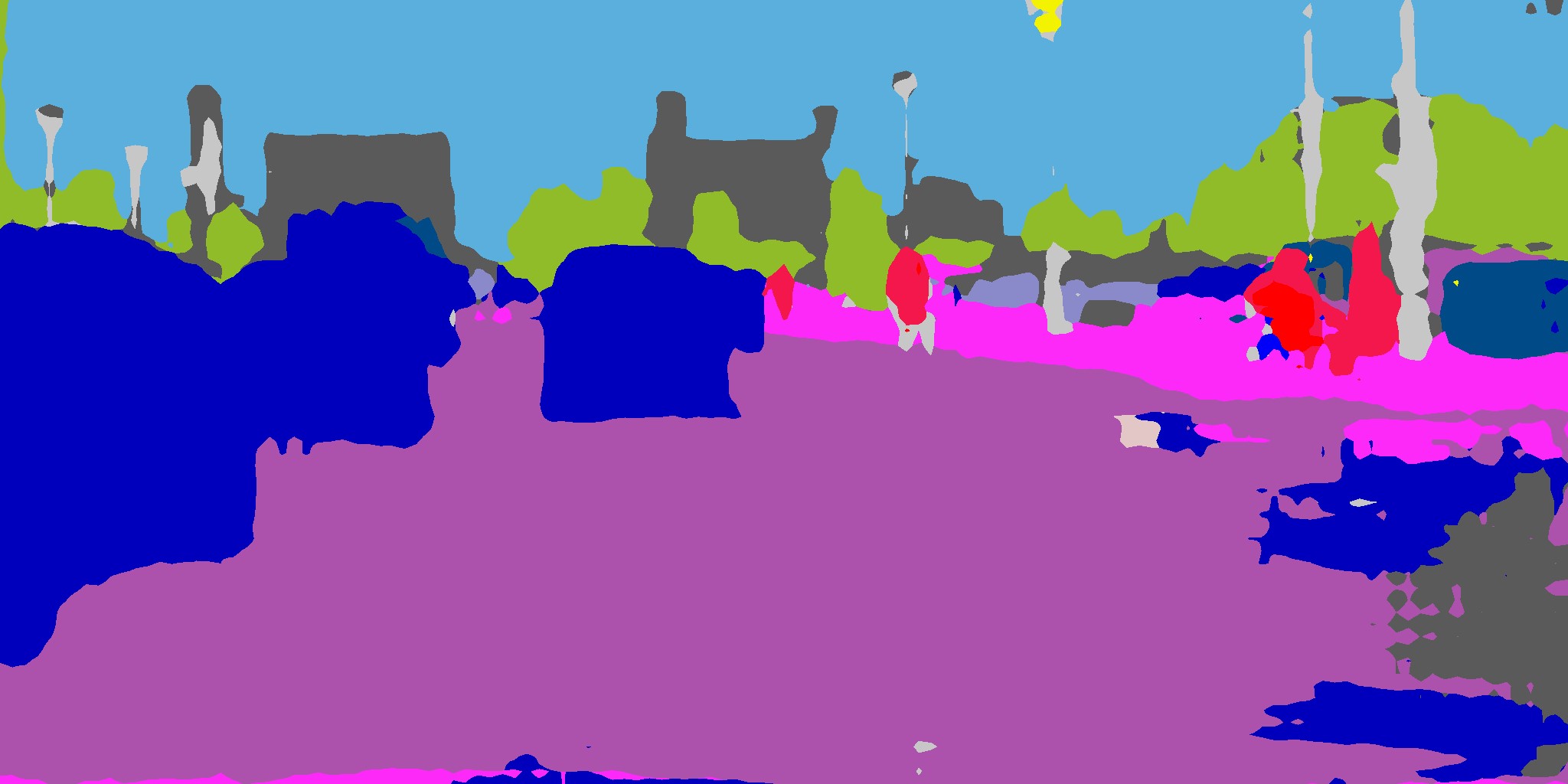}
		\end{subfigure}
		\begin{subfigure}[t]{0.24\textwidth}\centering
			\includegraphics[width=.98\textwidth]{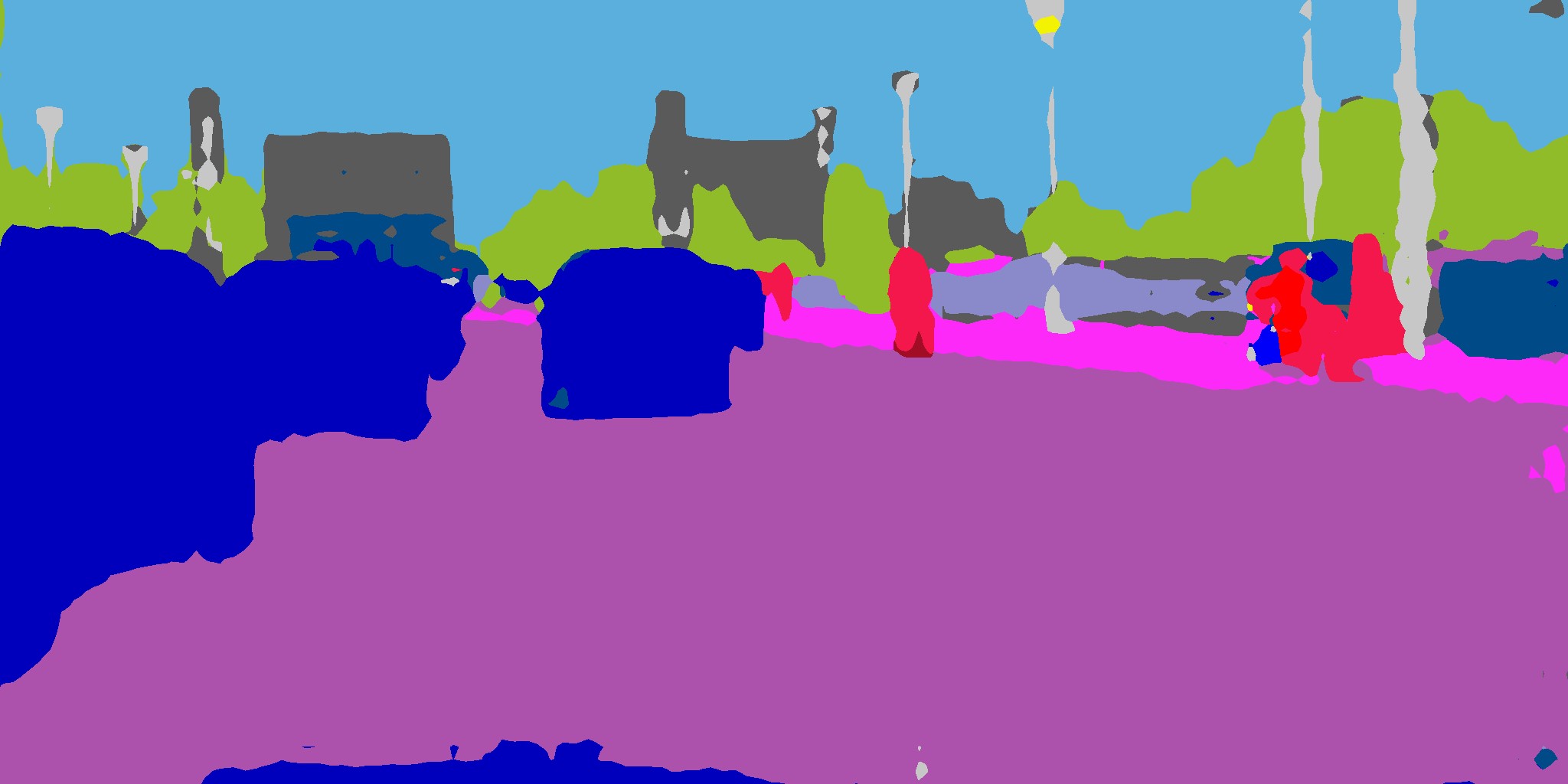}
		\end{subfigure}
	\end{center}
	\vspace{-0.3cm}
	\caption{\small \textbf{Qualitative results in the SYNTHIA$\rightarrow$Cityscapes (16 classes) set-up}. The four columns plot (a) RGB input images, (b) ground-truths, (c) AdvEnt baseline outputs and (d) DADA predictions. DADA shows good performance on `bus', `car', `bicycle' classes. Best viewed in color.}
	\vspace{-0.2cm}
	\label{fig:sup_qual_seg_1}
\end{figure*}

\vspace{-0.3cm}
\paragraph{SYNTHIA$\rightarrow$Cityscapes:}
In Table~\ref{lbl:tbl_res_16classes}, we report semantic segmentation performance in term of ``mean Intersection over Union'' (mIoU in \%) on the $16$ classes of the Cityscapes validation set.
DADA achieves state-of-the-art performance on the benchmark.
To the best of our knowledge, SPIGAN~\cite{lee2018spigan} is the only published work targeting the same problem that also considers depth as privileged information.
DADA achieves a D-Gain of $1.8\%$, almost double of SPIGAN's.
Analyzing per-class results, we observe that the improvement over AdvEnt \cite{vu2018advent} primarily comes from the `vehicle' category, \ie, `car' ($+7\%$), `bus' ($+8.1\%$) and `bike' ($+5.1\%$).
On `object' classes like `light' and `pole', DADA introduces moderate gains.
Figure~\ref{fig:sup_qual_seg_1} illustrates some qualitative examples comparing DADA and the AdvEnt baseline.
Our model shows better results on `vehicle' classes while AdvEnt sometimes makes severe mistakes of predicting `car' as `road' or `sidewalk'.
DADA also outperforms significantly other baseline methods that report results on a 13-class subset.

\begin{figure*}[t!]
	\begin{center}
		\begin{subfigure}[t]{0.24\textwidth}\centering
			\caption{Input image}\vspace{-0.2cm}
			\includegraphics[width=.98\textwidth]{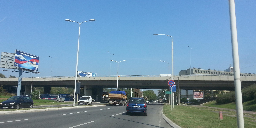}
		\end{subfigure}
		\begin{subfigure}[t]{0.24\textwidth}\centering
			\caption{GT}\vspace{-0.2cm}
			\includegraphics[width=.98\textwidth]{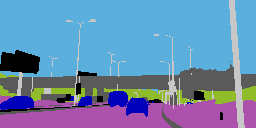}
		\end{subfigure}
		\begin{subfigure}[t]{0.24\textwidth}\centering
			\caption{SPIGAN}\vspace{-0.2cm}
			\includegraphics[width=.98\textwidth]{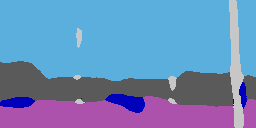}
		\end{subfigure}
		\begin{subfigure}[t]{0.24\textwidth}\centering
			\caption{DADA}\vspace{-0.2cm}
			\includegraphics[width=.98\textwidth, trim={40mm 60mm 162mm 45mm}, clip]{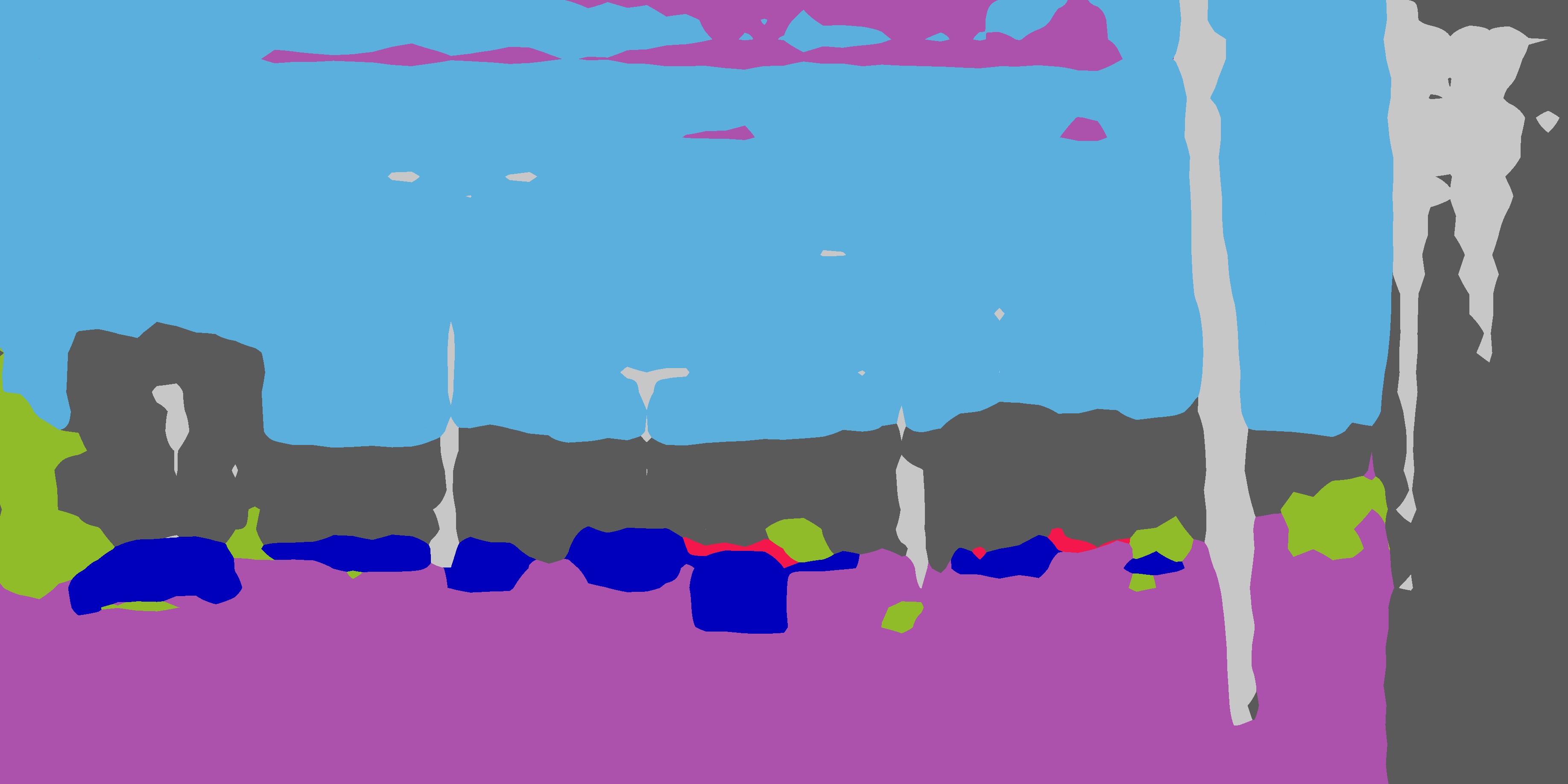}
		\end{subfigure}\\
		\vspace{0.02cm}
		\hdashrule[1ex][x]{17cm}{1.5pt}{1.5mm}\vspace{-0.13cm}
		\begin{subfigure}[t]{0.24\textwidth}\centering
			\includegraphics[width=.98\textwidth]{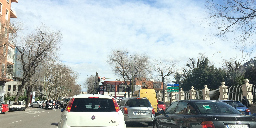}
		\end{subfigure}
		\begin{subfigure}[t]{0.24\textwidth}\centering
			\includegraphics[width=.98\textwidth]{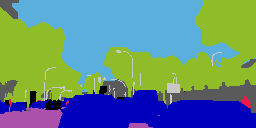}
		\end{subfigure}
		\begin{subfigure}[t]{0.24\textwidth}\centering
			\includegraphics[width=.98\textwidth]{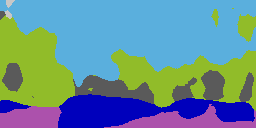}
		\end{subfigure}
		\begin{subfigure}[t]{0.24\textwidth}\centering
			\includegraphics[width=.98\textwidth, trim={43.605mm 130mm 476mm 130mm}, clip]{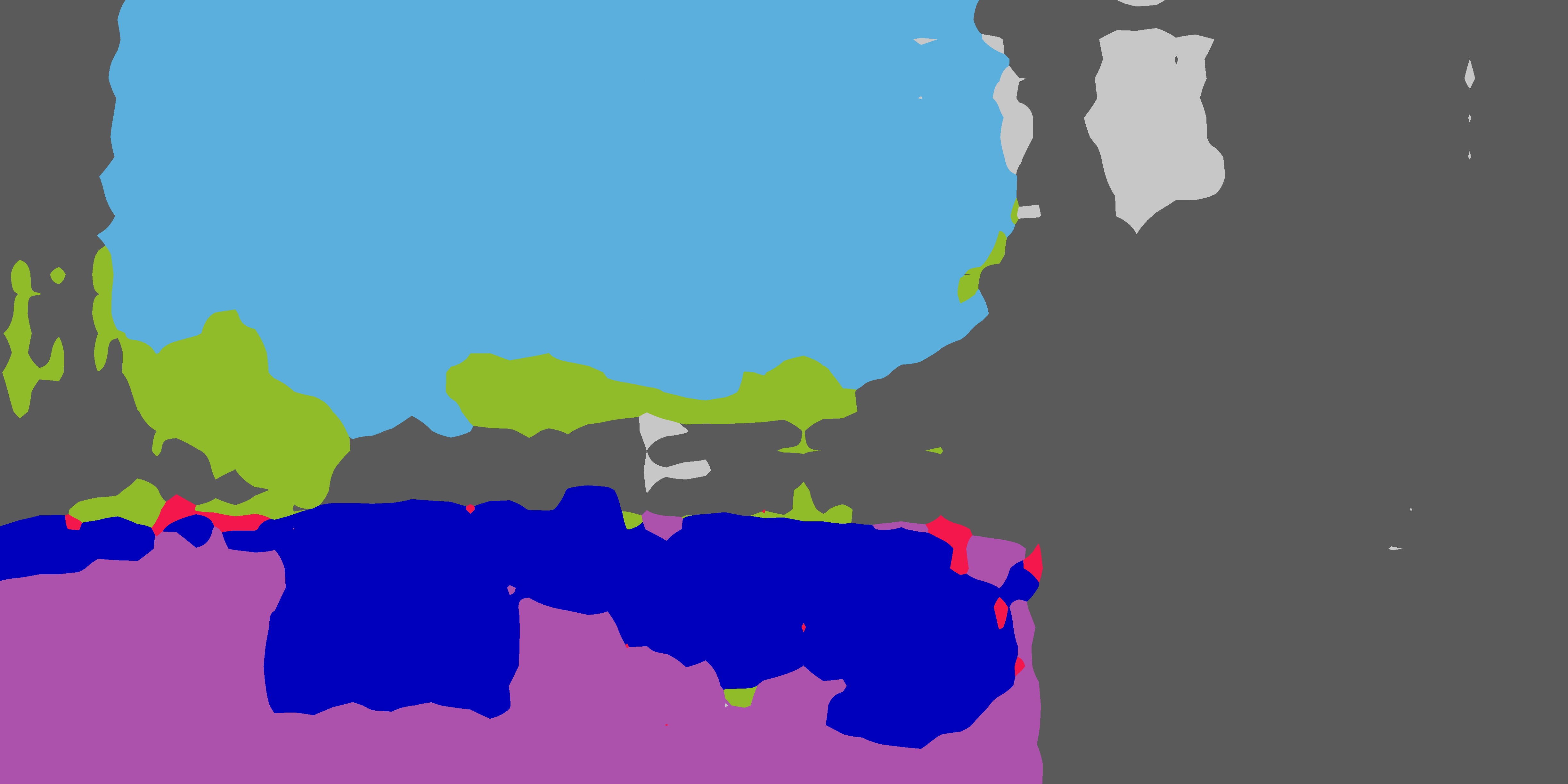}
		\end{subfigure}
	\end{center}
	\vspace{-0.3cm}
	\caption{\small \textbf{Qualitative results in the SYNTHIA$\rightarrow$Vistas (7 classes) set-up}. All models were trained and tested at the resolution of $320\times640$. From left to right, we show (a) RGB input images, (b) corresponding segmentation ground-truths, (c) SPIGAN's and (d) our DADA's segmentation predictions. Not only DADA performs visually better than SPIGAN, but it also produces correct predictions on wrongly annotated construction areas. Best viewed in color.}
	\label{fig:sup_qual_seg}
\end{figure*}

Table~\ref{lbl:tbl_res_7classes}-a shows results in the same experimental set-up, except that training and validation were done on $7$ categories.
To be comparable with~\cite{lee2018spigan}, we report additional results produced at the $320\times640$ resolution.
Our DADA framework outperforms the state-of-the-art on this benchmark by a large margin.
Over the AdvEnt baseline, we achieve $+3.2\%$ mIoU improvement.
Similar to the 16-class results, an important gain on the `vehicle' category ($+3.7\%$) is observed.
In addition, we contrast a $+3\%$ IoU on the `human' category to the negative IoU drops on classes `person' and `rider' reported in Table~\ref{lbl:tbl_res_16classes}.
We conjecture that this drop stems from the intra-category confusion, \ie, `pedestrian' and `rider' are easily confused.
A significant improvement is pointed out in the lower resolution set-up where using depth adds $+14.0\%$ to the `human' category IoU.
These results demonstrate the merit of DADA for UDA, especially on crucial categories like `human' and `vehicle' -- the vulnerable road users (VRUs).

An interesting UDA metric introduced in~\cite{lee2018spigan} is the negative transfer rate (lower is better) -- the percentage of after-adaptation test cases having per-image mIoUs lower than ones coming from the model trained only on source (without adaptation). On SYNTHIA$\rightarrow$Cityscapes (7 classes), $320\times640$ resolution DADA model has only $5\%$ negative transfer rate, compared to $9\%$ for SPIGAN.
It is worth noting that our only-on-source mIoU in this case is $50\%$, much larger than the one reported in~\cite{lee2018spigan} ($36.3\%$).
The AdvEnt baseline, without using depth, suffers from a negative transfer rate of $11\%$ -- more than double of DADA's.
\begin{table}[t]
	\begin{center}
		\begin{tabular}{l|cccc|c}
			\rule{0pt}{3ex}Setup &\rtb{ Surp. Adapt.}&\rtb{ Depth Adapt.}&\rtb{ Feat. Fusion}&\rtb{ DADA Fusion}&mIoU (\%)\\
			\hline
			\hline
			\rule{0pt}{3ex}S1 \small{(no adapt.)}&&&&&32.2 \\
			\hline
			\rule{0pt}{3ex}S2 \small{(AdvEnt)} &\checkmark&&&&40.8 \\
			\hline
			\rule{0pt}{3ex}S3&\checkmark&&\checkmark&&40.7 \\
			\hline
			\rule{0pt}{3ex}S4&&\checkmark&&&35.7 \\
			\hline
			\rule{0pt}{3ex}S5&&\checkmark&\checkmark&&38.0\\
			\hline
			\rule{0pt}{3ex}S6&\checkmark&\checkmark&\checkmark&&41.6\\
			\hline
			\rowcolor[gray]{.92}\rule{0pt}{3ex}S7 \small{(DADA)}&\checkmark&\checkmark&\checkmark&\checkmark&\textbf{42.6}
		\end{tabular}
	\end{center}
	\vspace{-0.3cm}
	\caption{\small \textbf{Segmentation performance (mIoU) on the Cityscapes validation set of $7$ ablation experiments}. Setup S1, with no check-marks, indicates source-only training.}
	\vspace{-0.3cm}
	\label{tbl:abl_dada}
\end{table}
\vspace{-0.3cm}
\paragraph{SYNTHIA$\rightarrow$Vistas:} In this experiment, the Mapillary Vistas~\cite{MVD2017} is used as the target domain. 
In SPIGAN~\cite{lee2018spigan}, the authors report unfavorable UDA behaviors on Vistas compared to Cityscapes. 
This seems caused by the artifacts that the source-target image translation introduces when trying to close the larger gap between SYNTHIA and Vistas.
In such a case, leveraging depth information demonstrates important adaptation improvement ($+17.3\%$).
On the other hand, our UDA framework does not undergo such a difficulty.
Indeed, as shown in Table~\ref{lbl:tbl_res_7classes}(a-b), the AdvEnt baseline performs much better than SPIGAN-no-PI, with no significant difference in absolute mIoU on the two target datasets ($59.4\%$ \textit{vs.} $54.0\%$).
Over such a stronger baseline, DADA still achieves an overall improvement of $+2.4\%$ mIoU.
We also obtain best per-class IoUs on the benchmark.

DADA (trained and tested on $320\times640$ images) has $30\%$ negative transfer rate compared to the $42\%$ of SPIGAN. As discussed in ~\cite{lee2018spigan}, the challenging domain gap between SYNTHIA and Vistas might cause  these high negative rates. In addition to this explanation, we also question the annotation quality of the Vistas dataset, visually inspection of results having revealed inconsistencies. 
Interestingly, when we evaluate the DADA model trained with the current set-up (SYNTHIA$\rightarrow$Vistas) on the Cityscapes validation set with arguably cleaner annotations, the obtained negative transfer rate reduces to $6\%$.

In Figure~\ref{fig:sup_qual_seg}, we show some qualitative results comparing our best model with SPIGAN.
As mentioned above, we note that the Vistas segmentation annotations are noisy.
For example, some construction areas slightly covered by tree branches are annotated as `vegetation'.
DADA provides reasonable predictions on these areas -- sometimes even better than human ground-truths.

\subsection{Ablation studies} \label{sec:exp_abl}
\paragraph{Effect of depth-aware adversarial adaptation.}
We report in Table~\ref{tbl:abl_dada} performance with seven training setups, S1 to S7: S1 is the source-only baseline (no  adaption at all), S2 amounts to AdvEnt (no use of depth) and S7 is DADA. Intermediate setups S3 to S6 amount to using or not the AdvEnt's ``surprisal adaptation'', the auxiliary depth adaptation (``depth adaptation''), the point-wise feature fusion mechanism (``feature fusion'') and the output-based fusion for DADA adversarial training (``DADA fusion'').
First, we remark that adversarial adaptation on the auxiliary depth-space (S4 and S5) does help improve performance of the main task.
Improvement of S5 over S4 demonstrates the advantage of depth fusion at the feature-level. Comparable performance of S2 and S3 indicates that depth supervision on the source domain is not effective in absence of depth-specific adaptation.
Indeed, S6, with two separate adversarial adaptations on the surprisal and depth spaces, works better than S2 and S3.
Still, in S6, the coupling between spaces remains loose as the adversarial losses are separately optimized.
Our depth-aware adaptation framework S7 employing both feature fusion and DADA fusion performs best: paying more attention to closer objects during adversarial training is beneficial.

\vspace{-0.3cm}
\paragraph{Annotation effort advantage.}
\begin{table}[t]
	\vspace{-0.2cm}
	\begin{center}
		\setlength{\tabcolsep}{5pt}
		\begin{tabular}{l|ccccc}
			\rule{0pt}{3ex}\% of SYNTHIA &10\%&30\%&50\%&70\%&100\%\\
			\hline
			\rule{0pt}{3ex}Cityscapes mIoU &32.6&35.1&40.9&41.0&\textbf{42.6}
		\end{tabular}
	\end{center}
	\vspace{-0.5cm}
	\caption{\small \textbf{DADA performance when trained on fractions of SYNTHIA}. Performance on Cityscapes as a function of the used percentage of training set.}
	\vspace{-0.4cm}
	\label{tbl:abl_percentage}
\end{table}
Table~\ref{tbl:abl_percentage} reports DADA's performance when trained on different fractions of the source dataset. Using only $50\%$ of the SYNTHIA images with segmentation and depth annotations, DADA achieves comparable performance to AdvEnt trained on all images with segmentation annotations ($40.9\%$ \vs $40.8\%$). This finding is of practical importance for real-world setups where the source domain is also composed of real scenes: while dense depth annotation remains automatic in this case (through stereo matching as in Cityscapes or densification of sparse LiDAR measurements), semantic annotation must be manual, which incurs high costs and quality problems. Annotating fewer scenes can thus be beneficial even if depth is additionally required.

\vspace{-0.3cm}
\paragraph{Limitations.}
We observe a few failure cases where different objects are indistinguishable due to blurry depth outputs.
Improving depth quality may help in such cases.
However, in our framework depth regression is only an auxiliary task which 
helps leveraging geometry-specific information to enrich visual representation and thus to improve on the main task.
As in~\cite{mordan2018revisiting}, paying too much attention to the auxiliary task actually hurts the performance on the main task.
	
	\section{Conclusion}
	In this work, we propose a novel UDA framework coined DADA -- Depth-Aware Domain Adaptation -- which leverages depth in source data as privileged information to help semantic  segmentation  on the target domain. This additional information is exploited through an auxiliary depth-prediction task that allows in turn a feature enrichment via fusion as well as a depth-aware modification of the original adaptation loss. Our experimental evaluations show that DADA consistently outperforms other UDA methods on different \textit{synthetic-2-real} semantic segmentation benchmarks. As a direction of future work, we envisage the extension to real-world scenarios where depth information in the source domain is only sparsely available, \eg, as provided by automotive laser scanners (LiDARs).

	{\small
		\bibliographystyle{ieee_fullname}
		\bibliography{egbib}
	}
	
\end{document}